\pdfoutput=1

\documentclass[11pt]{article}

\usepackage{EMNLP2022}

\usepackage{times}
\usepackage{latexsym}

\usepackage{graphicx}
\usepackage{subcaption,booktabs}
\usepackage{tabularx}
\usepackage{multirow, amsmath}
\usepackage{placeins}
\usepackage{gensymb} 
\usepackage{amssymb}

\newcommand{\blue}[1]{\textcolor{blue}{#1}}
\newcommand{\purple}[1]{\textcolor{purple}{#1}}

\usepackage{amsthm}

\usepackage[T1]{fontenc}

\usepackage[utf8]{inputenc}

\usepackage{microtype}

\usepackage{inconsolata}
\usepackage{CJKutf8}
\usepackage{textcomp}

\title{Counterfactual Recipe Generation: Exploring Compositional Generalization in a Realistic Scenario}

\author{
	Xiao Liu$^{1}$, 
	Yansong Feng$^{1,2}$\thanks{\;\;Corresponding author.}~~,
	Jizhi Tang$^{3}$,
	Chengang Hu$^{1}$\and
	Dongyan Zhao$^{1,4,5}$ \\
	$^1$Wangxuan Institute of Computer Technology, Peking University\\
	$^2$The MOE Key Laboratory of Computational Linguistics, Peking University\\
	$^3$Baidu Inc., Beijing, China\\
	$^4$Beijing Institute for General Artificial Intelligence\\
	$^5$State Key Laboratory of Media Convergence Production Technology and Systems\\
	{\tt \{lxlisa,fengyansong,hcg,zhaody\}@pku.edu.cn} \\
	{\tt tangjizhi@baidu.com}\\
}

\begin{document}
\maketitle
\begin{abstract}
People can acquire knowledge in an unsupervised manner by reading, and compose the knowledge to make novel combinations. In this paper, we investigate whether pretrained language models can perform compositional generalization in a realistic setting: recipe generation. We design the \emph{counterfactual recipe generation} task, which asks models to modify a base recipe according to the change of an ingredient. This task requires compositional generalization at two levels: the surface level of incorporating the new ingredient into the base recipe, and the deeper level of adjusting actions related to the changing ingredient. 
We collect a large-scale recipe dataset in Chinese for models to learn culinary knowledge, and a subset of action-level fine-grained annotations for evaluation.
We finetune pretrained language models on the recipe corpus, and use unsupervised counterfactual generation methods to generate modified recipes.
Results show that existing models have difficulties in modifying the ingredients while preserving the original text style, and often miss actions that need to be adjusted. Although pretrained language models can generate fluent recipe texts, they fail to truly learn and use the culinary knowledge in a compositional way. 
Code and data are available at \url{https://github.com/xxxiaol/counterfactual-recipe-generation}.
\end{abstract}
\section{Introduction}
Reading is an effective way to gain knowledge. When people read, mental processes like structured information extraction and rule discovery go on in our brains~\citep{gibson1975psychology}.
In the case of cooking, we read recipes of various dishes, gain knowledge of ingredients and flavors, and compose them to cook other dishes.~\footnote{We provide explanations of concepts related to recipes in Table~\ref{table-terminology} for better understanding.}

\begin{figure}
    \centering
    \includegraphics[width=0.9\columnwidth]{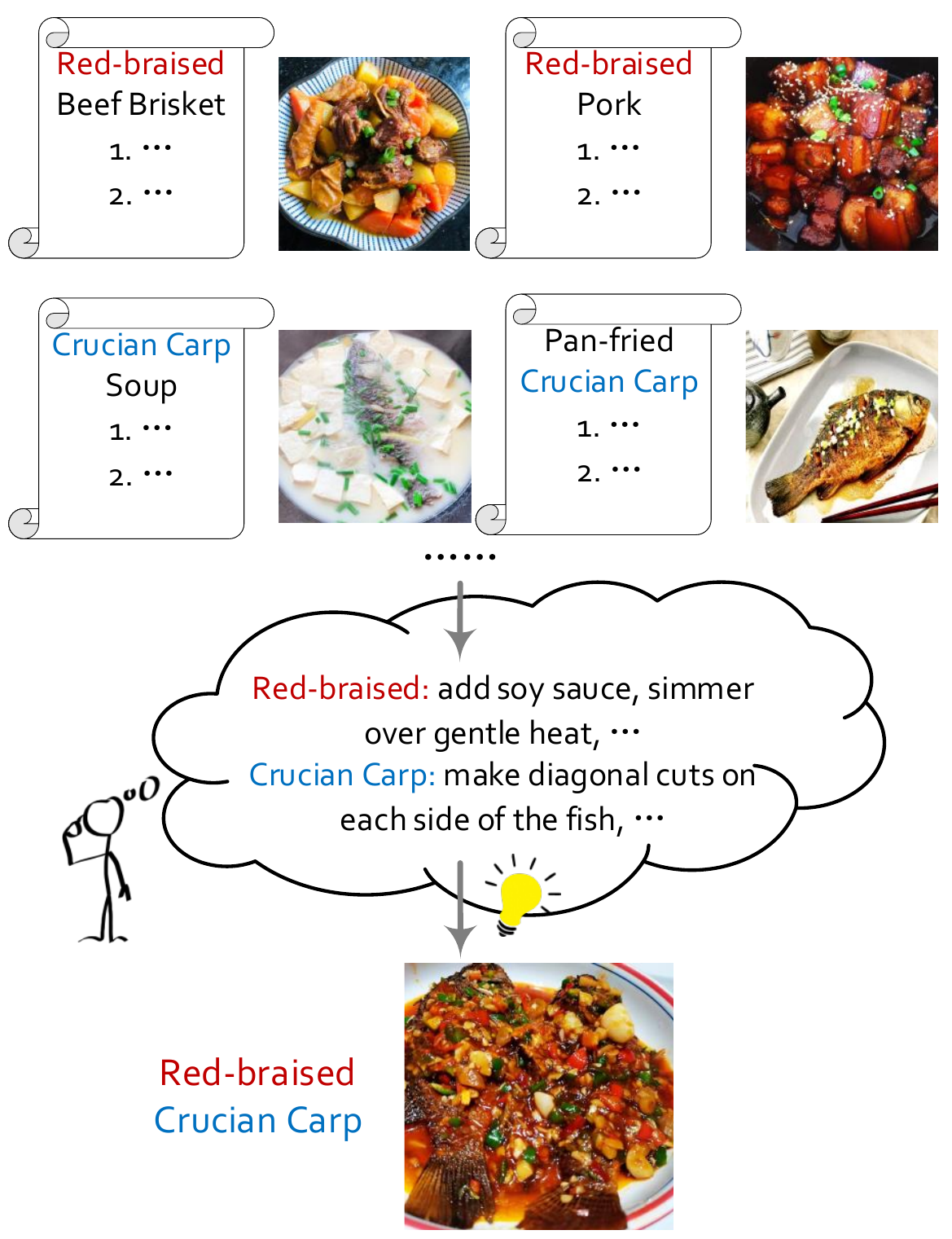}
    \caption{An example of the recipe learning process of human for the dish \emph{red-braised crucian carp}.}
    \label{fig-intro}
\end{figure}

This process involves knowledge acquisition and composition. As shown in Figure~\ref{fig-intro}, when people read recipes, they distill the knowledge of flavors and ingredients, like \emph{soy sauce is usually used to get red-braised flavor} and \emph{people often make diagonal cuts on fish to better marinate}. 
People can then cook new dishes like \emph{red-braised crucian carp} by composing existing knowledge about how to form the \emph{red-braised} flavor and how to cook \emph{fish}.

\begin{figure*}
    \centering
    \includegraphics[width=\textwidth]{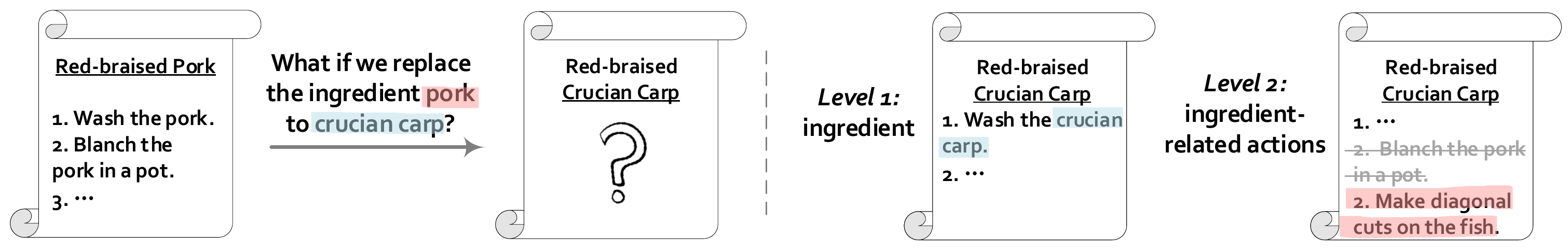}
    \caption{The counterfactual recipe generation task and the two levels of compositional competencies examined.} 
    \label{fig-task}
\end{figure*}
We expect models to acquire culinary knowledge unsupervisedly, and be able to use the knowledge skillfully, e.g., composing new dishes.
Current recipe processing tasks do not examine this ability explicitly. Recipe understanding tasks usually evaluate the models in a specific aspect under supervision, like identifying ingredient states or relationship between actions.
Recipe generation tasks evaluate whether models can generate fluent recipes, but do not investigate whether the generation ability relies on simple word correlation or culinary knowledge composition.

Current models exhibit strong compositional generalization ability in the understanding of synthetic texts~\citep{lake2019compositional,nye2020learning, weissenhorn2022compositional}, but few of them conduct experiments on realistic data, which is more challenging in two ways: (1) there are far many synonymous expressions in real-world texts, like various mentions of ingredients and actions in recipes~\citep{fang2022does}, but synthetic data often use a single expression for one meaning~\citep{keysers2019measuring}; (2) the knowledge in synthetic data is often expressed accurately and clearly, but knowledge in data provided by real users is varying. For example, when forming the \emph{red-braised} flavor, most people use soy sauce while few use iced sugar instead. Moreover, previous compositional generalization tasks mainly focus on semantic parsing and language grounding, while we aim to examine models in the form of natural language generation.

We propose the \textbf{Counterfactual Recipe Generation} task, to examine models' compositional generalization ability in understanding and generating recipe texts.
This task requires models to answer the question: given a recipe of a dish, how will the recipe change if we replace or add one ingredient? 
As shown in Figure~\ref{fig-task}, models are asked to modify the base recipe of \emph{red-braised pork} to form a new recipe of \emph{red-braised crucian carp} while preserving the original cooking and writing style.

We develop baseline methods to solve the task, as existing compositional generalization models cannot fit in the counterfactual generation task form. We finetune pretrained language models (PLMs) on our collected Chinese recipe corpus to learn culinary knowledge, and use prevalent unsupervised counterfactual generation frameworks to generate counterfactual recipes given \emph{\{base dish, base recipe, target dish\}}, where the target dish differs from the base dish in a main ingredient. 

Instead of annotating gold target recipes, we evaluate models in two levels of compositional competencies, which is less labor-consuming. 
The surface level (L1) is the fusion of the \textbf{changing ingredient} and the base recipe: the new ingredient should be added to the recipe, and the replaced ingredient should be removed. For instance, the original step \emph{wash the pork} needs to be changed to \emph{wash the crucian carp}. We evaluate the ingredient coverage ratio of the added and replaced ingredient, and the degree of recipe modification. Results show that existing PLMs can hardly cover the new ingredient and delete the replaced ingredient without making unnecessary modifications to the base recipe.

The deeper level (L2) is the fusion of \textbf{actions related to the changing ingredient} and the base recipe: actions to process the new ingredient should be inserted, and actions only related to the replaced ingredient should be deleted. This is a much harder problem involving decompositions and compositions of actions. In the case of Figure~\ref{fig-task}, a model should know \emph{blanch} is related to pork and is not suitable for a crucian carp, in order to delete the action \emph{blanch the pork}. Also, the action \emph{make diagonal cuts} is widely used in fish dishes, and should be added to the recipe. Our action-level evaluations show that most models fail to either remove all the irrelevant actions, or insert all the necessary ones, indicating that current PLMs have not fully learned the patterns of ingredient processing.

Our main contributions are as follows:
1) We propose the counterfactual recipe generation task to test models' compositional generalization ability in a realistic scenario.
2) We collect a large-scale Chinese recipe dataset and build a counterfactual recipe generation testbed with fine-grained action-level annotations, which can also increase the research diversity for procedural text understanding. 
3) We examine  models' compositional generalization ability from two levels. Our experiments show current PLMs are unable to modify the ingredient and preserve the original text style simultaneously, and will miss actions related to the changing ingredient that need to be adjusted.
Further analysis reveals that current models are still far from human experts in the deep understanding of procedural texts, like tracking entities and learning implicit patterns.

%

\begin{table*}[t]
    \centering
    \small
    \setlength{\tabcolsep}{4pt}
    \begin{tabular}{lll}
    \toprule
    \textbf{Word} & \textbf{Definition} & \textbf{Examples} \\
    \midrule
    Dish & Food prepared in a particular way. & \emph{Red-braised pork, Kung Pao chicken}\\
    Recipe & Instructions for preparing a dish. & \\
    Ingredient & Part of the foods that are combined to make a dish. & \emph{Pork, Chicken}\\
    Flavor & The taste expression of a dish. & \emph{Red-braised, Scorched chile} (the flavor of Kung Pao)\\
    Action & An event described in the recipe, centering on a verb. & \emph{Wash the pork, Blanch the pork}\\
    \bottomrule
    \end{tabular}
    \caption{Key concepts we used when analyzing recipes. }
    \label{table-terminology}
\end{table*}
\section{Related Work}
\subsection{Recipe Processing} 
Recipes are a common type of procedural texts, describing the actions a chef needs to perform to cook a specific dish. 
Recipe comprehension tasks include entity state tracking~\citep{bosselut2018simulating}, recipe structure extraction~\citep{kiddon2015mise, donatelli2021aligning}, and anaphora resolution~\citep{fang2022does}. These tasks involve the understanding of the relationship between ingredients and actions and test models' abilities under fine-grained supervision. In contrast, we evaluate whether models can understand the recipes and compose them unsupervisedly.
Recipe generation tasks ask models to create recipes from a given title. \citet{kiddon2016globally, h2020recipegpt} provide an ingredient list, \citet{majumder2019generating} add user's historical preference into consideration, and \citet{sakib2021evaluating} generate recipes from an action graph. 
\citet{li2021share} introduce the recipe editing task, which expects models to edit a base recipe to meet dietary constraints; and \citet{antognini2022assistive} iteratively rewrite recipes to satisfy users’ feedback. Our task is similar to the recipe editing tasks in including a base recipe in the input, but we go beyond simple ingredient substitution in recipes, and also evaluate whether the actions associated with the ingredients are alternated by models.


\subsection{Compositional Generalization}
Compositional generalization is the ability to understand and produce novel combinations of previously seen components and constructions~\citep{chomsky1956syntactic}.
To measure the compositional generalization ability of models, a line of research~\citep{lake2018generalization, ruis2020benchmark} designs tasks that map sentences into action sequences, and splits the data into training and testing sets from different data distributions, e.g., according to different lengths or primitive commands. 
Motivated by these works, we divide our data into finetuning corpus and test set based on flavors and ingredients, which are basic components of a dish.
Other compositional generalization works conduct experiments on semantic parsing~\citep{keysers2019measuring,kim2020cogs} and language grounding~\citep{johnson2017clevr}. To the best of our knowledge, our task is the first compositional generalization task in the form of natural language generation. 

Our task mainly differs from previous ones in measuring compositional generalization in a realistic setting, where all texts are natural rather than synthetic. The variation in natural language, both the variation of expressions and the variation of knowledge provided by different users, brings extra challenges. \citet{shaw2021compositional} also address the challenge of natural language variation in compositional generalization, but they experiment on semantic parsing, where the knowledge is highly consistent and can be inducted with grammar rules. 
\section{Task Definition}
We formulate our task in the form of counterfactual generation~\citep{qin2019counterfactual}:
\begin{equation}
    p(\mathbf{y}'|\mathbf{y}, \mathbf{x}_{\mathbf{y}}, \mathbf{x}'),
\end{equation}
where $\mathbf{y}$ in the base recipe, $\mathbf{x}_{\mathbf{y}}$ is the ingredient set of $\mathbf{y}$, $\mathbf{x}'$ is the adjusted ingredient set that replaces or adds one main ingredient, and $\mathbf{y}'$ is the target recipe to generate.
In Figure~\ref{fig-task}'s example, $\mathbf{y}$ is the base recipe of \emph{red-braised pork}, and $\mathbf{x}'$ differs from $\mathbf{x}_{\mathbf{y}}$ in changing \emph{pork} to \emph{crucian carp}.
\section{Data Preparation}
\subsection{The \textsc{XiaChuFang} Recipe Dataset}
We collect a novel Chinese dataset \textsc{XiaChuFang} of 1,550,151 recipes from \url{xiachufang.com}, a popular Chinese recipe sharing website. Compared to the commonly used English recipe dataset Recipe1M+, \textsc{XiaChuFang} contains 1.5 times the recipes. 
The website provides a list of common dishes. We map the recipe titles to these dishes, and find 1,242,206 recipes belonging to 30,060 dishes. A dish has 41.3 recipes on average. The average length of a recipe is 224 characters. Organizing recipes in terms of dishes helps us to learn what different people have in common when cooking this dish, which are often the necessary actions.

\begin{table}[t]
    \centering
    \small
    \setlength{\tabcolsep}{4pt}
    \begin{tabular}{ll}
    \toprule
    \textbf{Base Dish} & \textbf{Target Dish}\\
    \midrule
    Spicy \purple{Crayfish} & Spicy \blue{Chicken Feet}\\
    Kung Pao \purple{Chicken} & Kung Pao \blue{Shrimp Balls} \\
    Stir-fried \purple{Baby Cabbage} & Stir-fried \blue{Loofah} \\
    Cold Tofu & Cold Tofu with \blue{Century Egg} \\
    Fried Beef & Fried Beef with \blue{Carrot} \\
    \bottomrule
    \end{tabular}
    \caption{Examples of dish pairs used in evaluation.}
    \label{table-dish}
\end{table}
%
We select 50 dish pairs \emph{<$d_b$ (base dish), $d_t$ (target dish)>} for evaluation. The two dishes share the same flavor and differ in one principal ingredient in the dish name.~\footnote{Some auxiliary ingredients may also change accordingly, but we only focus on the changes directly associated with the principal ingredient changed in the dish name.}
We randomly sample 50 recipes of each base dish as the base recipes, and form 2,500 evaluation instances in total. By letting models modify different recipes for the same dish, we better measure whether models have actually learned the culinary knowledge and can apply the knowledge flexibly.

Table~\ref{table-dish} shows several dish pairs used in our evaluation, and the full list is in Appendix~\ref{appendix:dish_pairs}.
In the first three lines of Table~\ref{table-dish}, the target dish replaces one ingredient of the base dish; and in the last two lines, one ingredient is added to the target dish.
The latter situation is rare in the dataset (only 8\%), but it presents additional challenges, as simply substitution in recipes does not work, and models have to generate reasonable actions to add the new ingredient in the right place.

We regard the recipes that do not belong to the 50 dish pairs as the \emph{recipe corpus}. The corpus size is 1,479,764. We expect models to learn cooking knowledge unsupervisedly from this corpus.

The chosen dish pairs meet the following criteria: they are common in Chinese cuisine; recipes of the dishes have not been seen by the PLMs we used; and models have the opportunity to learn about the ingredients and flavors from the \emph{recipe corpus}. For example, for the ingredient \emph{crayfish}, models can learn how to process it from recipes of \emph{stir-fried crayfish}, \emph{garlic crayfish}, \emph{chilled crayfish}, etc. Details of the selection criteria are in Appendix~\ref{appendix:dish_pairs}.
\subsection{Pivot Actions}
\begin{figure*}
    \centering
    \includegraphics[width=\textwidth]{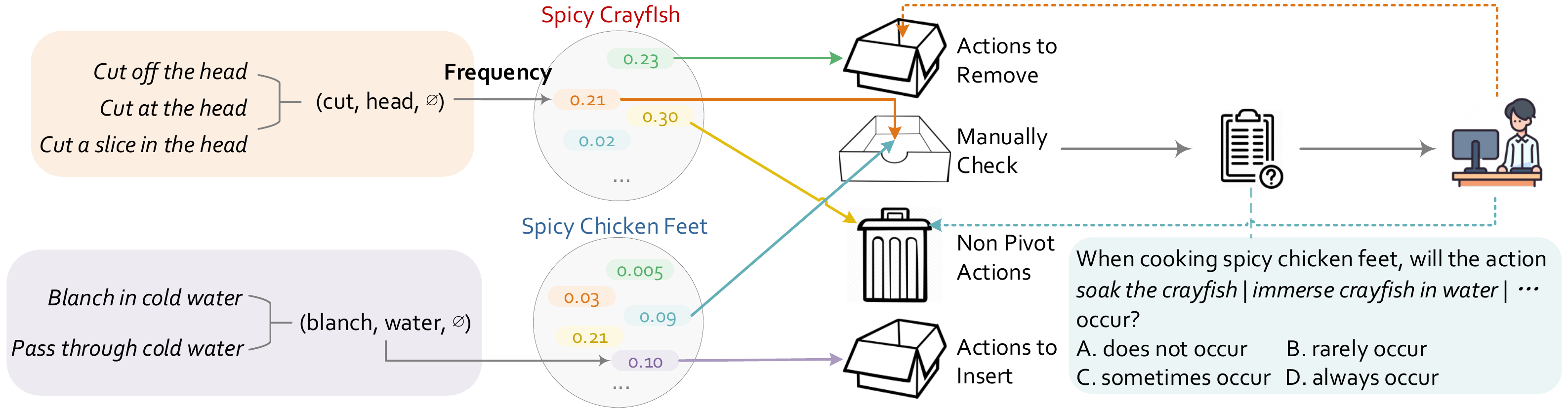}
    \caption{The process of collecting pivot actions.}
    \label{fig-pipeline}
\end{figure*}
Pivot actions are actions that differ between a dish pair, like \emph{blanch} and \emph{make diagonal cuts} in the case of \emph{<red-braised pork, red-braised crucian carp>}. Since there is no gold standard for the modified recipes, we evaluate the quality of the modified recipes by collecting the pivot actions.
For the dish pair \emph{<$d_b$, $d_t$>}, the pivot action set, $\mathcal{P}$, contains both actions to remove, $\mathcal{P}_R$, and actions to insert, $\mathcal{P}_I$.
$\mathcal{P}_R$ are actions that may appear in the recipes of $d_b$ but are \emph{not} appropriate for $d_t$; $\mathcal{P}_I$ are actions that are not needed for $d_b$ but \emph{should} appear in the recipes of $d_t$. In the example of Figure~\ref{fig-task}, \emph{blanch the pork} belongs to $\mathcal{P}_R$ and \emph{make diagonal cuts} belongs to $\mathcal{P}_I$. 

It is hard to ask annotators to write pivot actions from scratch, and checking all the actions in the recipes is inefficient. We observe that actions that frequently occur in the recipes of a dish are more likely to be necessary for the dish. Taking advantage of the abundant recipes of each dish in \textsc{XiaChuFang}, we categorize actions based on frequency, and ask annotators to annotate the vague ones. Figure~\ref{fig-pipeline} shows the semi-automatic pivot action collection workflow. 



\paragraph{Recipe Parsing.}
We first parse the recipe into a list of actions. An action $(v, igs, tools)$ is centered on a verb, often accompanied by ingredients and cooking tools. The definition is consistent with previous recipe processing works~\citep{donatelli2021aligning}, and the parsing details are in Appendix~\ref{appendix:parsing}. 

\paragraph{Pilot Study.}
\begin{figure}
    \centering
    \includegraphics[width=\columnwidth]{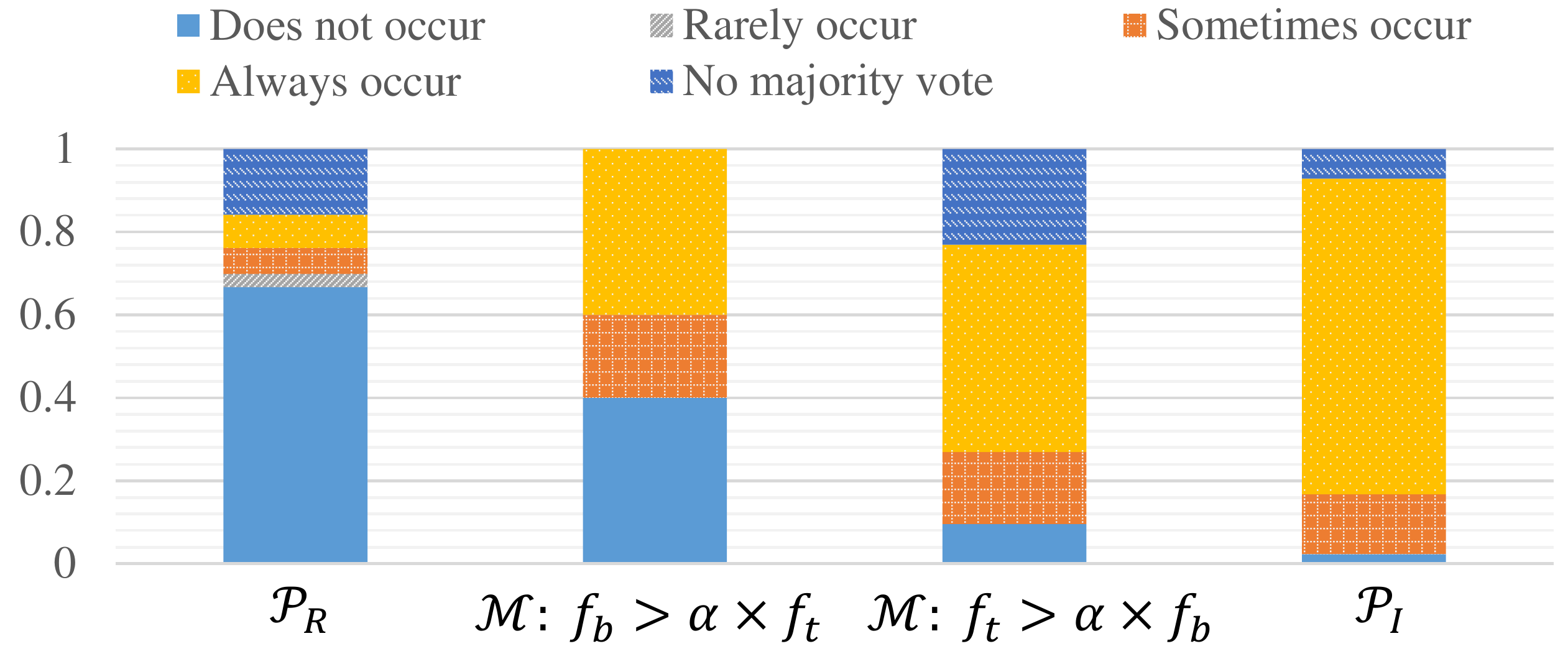}
    \caption{Pilot annotation results.}
    \label{fig-pilot}
\end{figure}
We conduct a pilot annotation to verify the validity of categorizing actions according to frequency. We randomly select 10 dish pairs to check whether annotators agree with the automatic categorizing results.

For each action $a$, we calculate its frequencies, $f_b$ and $f_t$, according to 
its appearances in the recipes of $d_b$ and $d_t$, respectively, and categorize it as:

$\mathcal{P}_R$ (actions to remove), if $f_b>\alpha \times f_t$ and $f_t<\tau_R$ (here $\alpha$ is a coefficient greater than 1), the criteria means that the action appears much often in recipes of $d_b$, and rarely appears in recipes of $d_t$;

$\mathcal{P}_I$ (actions to insert), if $f_t>\alpha \times f_b$ and $f_t>\tau_I$, which means that the action appears frequently in recipes of $d_t$, and rarely in recipes of $d_b$;

$\mathcal{M}$ (actions to manually check): if $a$ satisfies $f_b>\alpha \times f_t$ or $f_t>\alpha \times f_b$, but does not satisfy the threshold constraints ($f_t$ is not low or high enough).

We set $\alpha=5, \tau_R=0.01, \tau_I=0.1$ empirically, and ask annotators to annotate all actions in $\mathcal{P}_R \cup \mathcal{P}_I \cup \mathcal{M}$. We give annotators three instances of each action, and request them to determine whether the action \emph{does not occur, rarely occur, sometimes occur}, or \emph{always occur} in the target dish.

Annotation results are shown in Figure~\ref{fig-pilot}. We collect 3.4 answers for each question on average, and perform a majority vote (more than half of the annotators agree with it). 
More than 2/3 actions in $\mathcal{P}_R$ are annotated as \emph{do not occur}, and more than 2/3 actions in $\mathcal{P}_I$ are annotated as \emph{always occur}. After a careful check over those disagreement cases (in Appendix~\ref{appendix:disagreement}), we find that the disagreements are mainly due to the subjectivity of cooking.
Actions in $\mathcal{P}_R$ are more likely to be annotated as \emph{do not occur}, and actions in $\mathcal{P}_I$ are more likely to be annotated as \emph{always occur}, comparing to actions in $\mathcal{M}$. These indicate the validity of distinguishing the automatically mined actions ($\mathcal{P}_R$ and $\mathcal{P}_I$) and those remaining to be manually checked. 

\paragraph{Human Annotation.}
We then perform human annotation on the set $\mathcal{M}$ of all dish pairs. Each question is annotated by at least three annotators. All annotators have more than one year of culinary experience, and a questionnaire sample is in Appendix~\ref{appendix:annotation}. 

We add actions with the majority vote \emph{do not occur} into $\mathcal{P}_R$, actions with the majority vote \emph{always occur} into $\mathcal{P}_I$, and discard the others.
A dish pair has 22.4 pivot actions on average: 13.6 to remove and 8.8 to insert.

\subsection{Action Location}
The order of actions is not arbitrary, as actions affect ingredient states such as shape and temperature~\citep{bosselut2018simulating}. Therefore, models are expected to not only generate the necessary actions, but also locate them in the right place. 
We judge the correctness of the position according to the order constraints, i.e., examining which actions should appear before or after the inserted action.
That is, for each action $a$ to insert, it should be placed after actions $\mathcal{A}_p^a$ and before actions $\mathcal{A}_s^a$. The order constraints can be seen as causal dependencies~\citep{pan2020multi}: actions in $\mathcal{A}_p^a$ are causes of $a$, and actions in $\mathcal{A}_s^a$ are effects of $a$. 

Previous works use pre-defined action graph or train a classifier to recognize the relationship between actions~\citep{mori2014flow,pan2020multi}. Thanks to the plentiful recipes of each dish in our dataset, we are able to identify the causal dependencies between actions using causal estimation techniques.
We regard the pivot action $a$ as the \emph{effect}, and exhaustively take every other action $a'$ that appears before $a$ in the recipes of $d_t$ as the \emph{cause}. All other actions that appear before $a$ and $a'$ are considered as potential confounders. We estimate the averaged causal effect $\psi_{a', a}$ with propensity score matching~\citep{rosenbaum1983central}. If $\psi_{a', a}>\tau_c$, 
we put $a'$ into $\mathcal{A}_p^a$. We set $\tau_c=0.1$ empirically, and construct $\mathcal{A}_s^a$ in the same way: setting $a$ as the cause and other actions as the effect.
\section{Baseline Methods}
We use two competitive unsupervised counterfactual generation  models~\citep{qin2020back,chen2021unsupervised} as our baselines. We do not consider existing compositional generalization models as it is hard to adapt them to our task setup. In order to make those models learn culinary knowledge, we finetune GPT-2 on the recipe corpus and use it as the backbone of the baselines.
\subsection{Learn from Recipes}
\label{sec:finetune}
There are various ways to exploit the recipe corpus. We take a common practice in our baseline method: following the pretraining objective to finetune the model. 
Specifically, we use GPT-2~\citep{radford2019language} pretrained on CLUECorpus2020~\citep{xu2020cluecorpus2020,zhao2019uer}. We concatenate the dish name and the recipe as the input: \emph{[Dish Name] is made as follows. [Recipe Text].}, and finetune GPT-2 with the causal language modeling objective: predicting the next word given all the previous ones. We finetune GPT-2 with learning rate 5e-5 for 10 epochs. 

\subsection{Perform Counterfactual Generation}
Existing unsupervised counterfactual generation methods use off-the-shelf PLMs with no extra supervision, and perform well on the TimeTravel~\citep{qin2019counterfactual} dataset. Among them, \textsc{EduCat}~\citep{chen2021unsupervised} and D\textsc{elorean}~\citep{qin2020back} are two representative works and they differ in the modification style: \textsc{EduCat} edits locally and explicitly, while \textsc{Delorean} refines the text implicitly from a holistic view.
Besides the two models, we also take two ways to directly utilize GPT-2 for comparison. 
%

\textbf{GPT-2 (D)}\quad In the basic setting, we ask GPT-2 to predict after \emph{[Target Dish] is made as follows.} without considering the base recipe. This is consistent with the finetuning setting, and can exhibit the model's recipe generation ability without being affected by the base recipe. 

\textbf{GPT-2 (D+R)}\quad  We try to input the task instruction and the base recipe into GPT-2: \emph{Please rewrite the recipe for [Target Dish] based on the recipe for [Base Dish]. [Base Recipe]. [Target Dish] is made as follows.} This setting tests whether the model can understand the instruction and compose its knowledge of the target dish into the base recipe.

\textbf{\textsc{EduCat}} is an editing-based model. It starts from the base recipe and edits step by step to meet the target dish. In each step, \textsc{EduCat} detects possible positions in the current generated recipe contradictory to the target dish, proposes a word-level modification operation (\textit{replace/insert/delete} at the position), and decides whether to accept the modification according to its fluency and coherence scores. 

\textbf{\textsc{Delorean}} is a constraint-based model. It generates based on the target dish first and iteratively refines the generated recipe to meet the two constraints: the recipe should be coherent with the target dish while remaining similar to the base recipe. \textsc{Delorean} measures the similarity with the KL divergence between the generated recipe and the base recipe.

We equip \textsc{Delorean} and \textsc{EduCat} with the finetuned GPT-2, and use their original hyperparameters.


\section{Evaluation}
\begin{table*}[t]
    \centering
    \small
    \begin{subtable}[t]{0.47\textwidth}
        \centering
        \setlength{\tabcolsep}{3.5pt}
        \begin{tabular}{lcccc}
        \toprule
        & \multicolumn{2}{c}{\textbf{CoI}} & \multicolumn{2}{c}{\textbf{Preservation}}\\
        \cmidrule(lr){2-3}
        \cmidrule(lr){4-5}
        & \textbf{Add} & \textbf{Replace}$\downarrow$ & \textbf{BLEU} & \textbf{BERTScore} \\
        \midrule
        GPT-2 (D) & 74.0 & \textbf{0.0} & 1.4 & 68.5 \\
        GPT-2 (D+R) & 71.8 & 15.2 & 3.9 & 70.6 \\
        \textsc{EduCat} & 43.5 & 51.3 & 24.4 & 81.4\\
        \textsc{Delorean} & \textbf{90.8} & 0.3 & 3.0 & 67.4 \\
        \bottomrule
        \end{tabular}
        \caption{L1: coverage of ingredients (CoI) and extent of preservation of different models. $\downarrow$ means smaller is better. The preservation scores can not be directly compared (see Section~\ref{sec:l1_results}).}
        \label{table-preliminary}
    \end{subtable}
    \hspace{0.5em}
    \begin{subtable}[t]{0.51\textwidth}
        \centering
        \setlength{\tabcolsep}{3.1pt}
        \begin{tabular}{lcccccccc}
        \toprule
        & \multicolumn{3}{c}{\textbf{Hard}} & \multicolumn{5}{c}{\textbf{Soft}}\\
        \cmidrule(lr){2-4}
        \cmidrule(lr){5-9} 
        & \textbf{P} & \textbf{R} & $\mathbf{F_1}$ & \textbf{P} & \textbf{R} & $\mathbf{F_1}$ & $\mathbf{F_1^+}$ & $\mathbf{F_1^-}$\\
        \midrule
        GPT-2 (D) & 11.8 & 33.8 & 15.9 & 21.7 & \textbf{56.8} & \textbf{29.0} & \textbf{27.7} & 31.0 \\
        GPT-2 (D+R) & 13.1 & 34.3 & 17.2 & 21.4 & 53.2 & 27.9 & 25.8 & 31.2 \\
        \textsc{EduCat} & 8.9 & 16.2 & 10.0 & 17.7 & 29.1 & 19.5 & 11.5 & 27.8 \\
        \textsc{Delorean} & \textbf{13.5} & \textbf{34.6} & \textbf{17.5} & \textbf{22.4} & 52.5 & 28.6 & 26.5 & \textbf{31.6} \\
        \bottomrule
        \end{tabular}
        \caption{L2: action-level evaluation. $F_1^+$ and $F_1^-$ indicate $F_1$ on actions to insert and actions to remove respectively.}
        \label{table-action}
    \end{subtable}
    \caption{Evaluation results of two compositional competency levels.}
\end{table*}
\subsection{L1: Surface-level Evaluation}
We first evaluate the generated texts from two aspects, to answer the L1 question: Can models incorporate the changing ingredient into the base recipe without making excessive changes?
\subsubsection{Evaluation Metrics}

\paragraph{Coverage of Ingredients.}
We check the coverage ratio of newly added ingredients in the generated recipes. For instances where an old ingredient is replaced, we also check whether the replaced ingredient still exists in the generated recipes:
\begin{equation}
    CoI=\frac{\sum_{i=1}^{N} \textbf{1}[ing_i \in r_i]}{N},
\end{equation}
where $ing_i$ is the added/replaced ingredient in the $i$-th target dish , and $r_i$ is the generated recipe. We expect the coverage ratio of the added ingredient to be high, and the coverage ratio of the replaced ingredient to be low.
The metric only captures if the ingredient appears in the recipe, but does not check whether the ingredient is processed correctly, which will be examined in the action-level evaluation.

\paragraph{Extent of Preservation.}
We calculate the correspondence between the generated recipe and the base recipe with BLEU~\citep{papineni2002bleu} and BERTScore~\citep{bert-score}.
The correspondence is not expected to be 100\% as one major ingredient changes, but we expect the score not to be too low, as it indicates better preservation. 

\subsubsection{Results}
\label{sec:l1_results}
From Table~\ref{table-preliminary}, we observe that \textsc{Delorean} successfully covers most of the added ingredients and removes most of the replaced ingredients, benefiting from its decoding strategy that starts from the target dish and pushes the generated text to be coherent with the target dish in the whole generation. 
GPT-2 (D) exhibits the generation ability of finetuned PLMs. Without the disturbance of the base recipe, it will not generate the replaced ingredient, but it still misses the added ingredient in several cases. 
By adding the task instruction and the base recipe into the input, GPT-2 (D+R) underperforms GPT-2 (D) on both added and replaced ingredients, indicating the model may not be able to fully understand the instruction and is interfered by the base recipe.
\textsc{EduCat} does not perform well in ingredient coverage, failing to insert many ingredients in the target dish and still keeping many replaced ingredients. It makes many unnecessary modifications but fails to detect the true contradictory between the target dish and the recipe. 

On the other hand, it is difficult for all models to preserve the style of the base recipe. The BLEU scores of GPT-2 (D), GPT-2 (D+R), and \textsc{Delorean} are less than 5\%. As a reference for assessment, we ask human experts to conduct the counterfactual writing task (the recipes they write are further used in the human evaluation), and the BLEU score between expert-written recipes and the base recipes is 65.2\%. Compared with human experts, only \textsc{EduCat} well preserves the original style, and the other models perform poorly in preservation.
Although \textsc{Delorean} constrains the generated text to be similar to the base recipe, the constraint is still weak and it actually modifies most of the base recipes, as its correspondence with the base recipe is on par with GPT-2 (D). 

These results show that existing models still have difficulties in the basic composition. They cannot decompose the replaced ingredient or compose the added ingredient into the recipe well, indicating the deficiency in identifying and tracking an ingredient. 

\subsection{L2: Action-level Evaluation}
We design action-level evaluation metrics to answer the L2 question: Can models identify pivot actions when the ingredient changes, remove the inappropriate actions and compose the necessary actions into the base recipe? 

\subsubsection{Evaluation Metrics}
We measure the precision, recall, and $F_1$ score between the changes, $\mathcal{C}$, made by a model and the pivot actions, $\mathcal{P}$. $\mathcal{C}$ is constructed by parsing both the base and generated recipes, and finding the different actions. For inserted actions in $\mathcal{C}\cap\mathcal{P}_I$, we additionally examine their positions in the generated recipe, and only those that meet the order constraints are kept in the intersection.

Considering different expressions may have the same or similar meaning, we also design a \emph{soft} version measurement (and those in the above paragraph are regarded as \emph{hard}). In the soft version, we add actions of similar surface forms $sim(ph(a), ph(a'))>\tau_s$ into the intersection, using the similarity score as their weight (the weight of actions in the hard intersection is always 1).
$ph(a)$ is the text phrase corresponding to action $a$, $a \in \mathcal{C}$ and $a' \in \mathcal{P}$. We calculate $sim(\cdot)$ with cosine similarity of averaged word embeddings, and set $\tau_s$ to 0.9. 

\subsubsection{Results}
From Table~\ref{table-action}, we find that the action-level performance is quite low: all models achieve no more than 20\% hard $F_1$ and no more than 30\% soft $F_1$.
\textsc{EduCat} performs the worst, probably because it modifies the base recipe word by word. Even if the finetuned PLM learns certain pivot actions, it is still hard for \textsc{EduCat} to add or delete the entire action as it may need several modification steps.
Comparing with the editing-based method, \textsc{Delorean} performs relatively better, as it considers the generated recipe and the new ingredient from a holistic perspective. This makes it easier to find the actions that are consistent or contradictory with the target dish, and adds/removes an action at once. 
The performance of \textsc{Delorean} and the two ways of directly using GPT-2 are similar, as \textsc{Delorean} changes most of the base recipe.

Inserting necessary actions is harder than removing inappropriate actions for all models. Insertion requires models to link unseen actions with the ingredients, and locate the actions in right places.
GPT-2 (D) locates almost all of the inserted actions correctly 
(99\% of them meet the order constraints). The order accuracy is 87\% for GPT-2 (D+R), 84\% for \textsc{EduCat} and 80\% for \textsc{Delorean}. Compared with generation from scratch, the difficulty of action positioning raises when models are asked to rewrite on the base recipe. Models learn the order of actions from the finetuning recipe corpus with the causal language modeling objective, so it is more natural to generate actions sequentially. 
\section{Analysis}
\subsection{Human Evaluation}
We conduct a human evaluation to assess model performance more comprehensively, and to check whether our evaluation results are consistent with human impressions. 
We randomly select one instance for each dish pair, and ask raters to choose the best and worst generated recipes on \emph{grammaticality} (whether the text is grammatically correct and fluent), \emph{correctness} (whether the text is a correct recipe of the target dish), and \emph{preservation} (whether the text preserves the base recipe's cooking and writing style).

Expert-written recipes are also added into comparison (referred to as \emph{Expert}). 
We ask cooking experts, who have cooked for more than a decade, to write counterfactual recipes for the 50 base recipes. As GPT-2 (D) and GPT-2 (D+R) perform comparably in previous evaluations, we only keep GPT-2 (D) as a baseline of directly using the PLM.

We obtain the ratings with Best-Worst Scaling~\cite[BWS,][]{louviere1991best}: the percentage of a model being selected as the best minus the percentage of being selected as the worst. The score ranges from -100 to 100, and the higher is better. BWS is shown to be more reliable than Likert scales and is used in previous NLP human evaluations~\citep{kiritchenko2017best}. Each recipe is rated by at least two people.

\begin{table}[t]
    \centering
    \small
    \setlength{\tabcolsep}{4pt}
    \begin{tabular}{lrrr}
    \toprule
    & \textbf{Grammar} & \textbf{Correctness} & \textbf{Preservation} \\
    \midrule
    GPT-2 (D) & -30.6 & -7.3 & -72.6 \\
    \textsc{EduCat} & -1.0 & -63.1 & 13.8 \\
    \textsc{Delorean} & -44.1 & -12.0 & -18.9 \\
    Expert & \textbf{75.6} & \textbf{82.5} & \textbf{77.8} \\ 
    \bottomrule
    \end{tabular}
    \caption{Human evaluation results.}
    \label{table-human}
\end{table}
Results are shown in Table~\ref{table-human}. 
\emph{Expert} is much better than all three models in all three aspects, indicating there is a big gap between the counterfactual recipe generation capabilities of models and humans. 
Among the three models, GPT-2 (D) is better in the \emph{correctness} of cooking, which is consistent with the soft action-level evaluation results. Based on the same PLM, models' performance hurt when they try to go closer to the style of the base recipe.
\textsc{EduCat} is better in the \emph{preservation} of the base recipe as it modifies word by word on the base recipe. This is also consistent with the preliminary preservation results. 
\textsc{Delorean} is in the middle regarding \emph{correctness} and \emph{preservation}, but performs the worst in \emph{grammaticality}. Considering its high ingredient coverage ratio, it achieves surface-level composition of the new ingredient and the base recipe, but the composition is rigid and incoherent.
We also conduct case studies analyzing model mistakes in Appendix~\ref{appendix:case}.

\subsection{Challenges} 
We discuss challenges of learning culinary knowledge and composing them in the recipe scenario.

\paragraph{Various mentions of ingredients.} An ingredient may have different names in Chinese recipes, like \begin{CJK*}{UTF8}{gbsn}鸡爪\end{CJK*} (chicken feet) is also called \begin{CJK*}{UTF8}{gbsn}鸡脚\end{CJK*} or \begin{CJK*}{UTF8}{gbsn}凤爪\end{CJK*}. Moreover, the mentions of ingredients may change when their status change, like \begin{CJK*}{UTF8}{gbsn}猪肉\end{CJK*} (pork) converts to \begin{CJK*}{UTF8}{gbsn}肉丝\end{CJK*} (shredded pork) after cut. We sample 20 recipes from the recipe corpus, and find the variation of mentions is quite common: the main ingredient has 2.1 different mentions on average. This adds to the difficulty of identifying and tracking ingredients, which is involved in the L1 evaluation.

\paragraph{Omission of ingredients in recipes.} Object omission is commonly seen in Chinese~\citep{huang1998logical}, especially in informal writings like recipes. For example, the object ingredient \emph{squid} appears only once at the beginning in the sequence of actions \begin{CJK*}{UTF8}{gbsn}洗净鱿鱼，去除中间墨囊，切成块。焯过捞起沥水。\end{CJK*} (\emph{Wash the squid, remove the middle ink sac, cut into pieces. Blanch and drain.}) 
Each sampled recipe has an average of 3.3 actions where the ingredient is omitted, which increases the difficulty of linking actions with ingredients and may affect the L2 performance.

\paragraph{Variety of cooking styles.} Different people have different cooking styles, so the recipes they post for the same dish also vary. For example, when cooking \emph{stir-fried shredded potatoes}, some people \emph{blanch} potatoes to make them crispier, while others may prefer the soft texture. Models may be confused by the varying cooking ways, and this poses new challenge in compositional generalization.

\paragraph{Order determination.} As recipes are sequential and order-sensitive, models need to find the appropriate location when inserting an action. Learning the order knowledge is not trivial, which requires models to separate \emph{sequential} actions from \emph{parallel} ones and find the cause-and-effect relationship between actions.




\section{Conclusion}
We propose counterfactual recipe generation, a task to evaluate models' compositional generalization ability in realistic settings. We explore two levels of compositional competencies: modifying the changing ingredient, and modifying the ingredient-related actions in the base recipe. Experimental results show that current models have difficulties in composing the new ingredient and actions into the base recipe while preserving the cooking and writing styles. By proposing this task, we reveal the large gap between current pretrained language models and humans in the deep understanding of procedural text, and want to elicit attention to pretrained language models' abilities in unsupervisedly learning and generalizing in real-life scenarios.

\section*{Limitations}



\paragraph{Scalability to datasets of other languages.} In this paper, we only consider common Chinese dishes which merely take up a small portion of those across the world. Our counterfactual recipe generation task can be extended to datasets of other languages and other cuisines, and it is promising to take the geo-diversity~\cite{yin2021broaden} of cooking knowledge into account. For example, \emph{<beef stew, venison stew>} is a potential dish pair in Irish cuisine, and the pivot actions of \emph{venison stew} 
are different from Chinese cooking. At the practical level, we make use of various recipes of each dish in pivot action collection and action location. So the dataset needs to contain a certain number of recipes for each evaluated dish, in order to measure the necessities of actions with their frequencies.

\paragraph{Alignment of multiple surface forms.} There are various expressions that have the same meaning in human-written recipes. We use soft action-level measurements to tackle the problem, but more human involvement may be needed to align the expressions accurately. This also limits the evaluation of ingredient coverage. We only consider the ingredient name that appears in the dish name, as we are unable to take all mentions of the ingredient into consideration. 

\paragraph{Imperfection of evaluation metrics.} It is hard to precisely evaluate generated text in natural language generation tasks, and our task also faces this challenge. We only consider some aspects to evaluate the counterfactually generated recipe, including the coverage of ingredients, the correspondence with the base recipe, and the coverage of pivot actions. Other aspects like the discourse structure of the recipe can also be taken into account.

\paragraph{Lack of corresponding solutions.} Although we show that current models' compositional generalization abilities are relatively poor in the recipe generation scenario, we did not propose solutions for improvement. Facing the challenges we discussed, we intend to design corresponding methods in the future.
\section*{Ethics Statement}
\paragraph{Intellectual Property.} We ensure that intellectual property of the original authors of recipes in \textsc{XiaChuFang} is respected during data collection with permission of licence\footnote{\url{https://www.xiachufang.com/principle/}}. And the collected data would not be used
commercially.


\section*{Acknowledgments}
This work is supported in part by NSFC (62161160339) and National Key R\&D Program of China (No. 2020AAA0106600). We would like to thank the anonymous reviewers for the helpful suggestions, and our great annotators for their careful work, especially Yujing Han, Yongning Dai, Meng Zhang, Yanhong Bai, Li Ju, Siying Li, and Jie Feng.  Also, we should thank Chen Zhang, Da Yin, Quzhe Huang and Nan Hu for their detailed comments. For any correspondence, please contact Yansong Feng.

\bibliography{custom}

\begin{thebibliography}{37}
\expandafter\ifx\csname natexlab\endcsname\relax\def\natexlab#1{#1}\fi

\bibitem[{Antognini et~al.(2022)Antognini, Li, Faltings, and
  McAuley}]{antognini2022assistive}
Diego Antognini, Shuyang Li, Boi Faltings, and Julian McAuley. 2022.
\newblock Assistive recipe editing through critiquing.
\newblock \emph{arXiv preprint arXiv:2205.02454}.

\bibitem[{Bosselut et~al.(2018)Bosselut, Levy, Holtzman, Ennis, Fox, and
  Choi}]{bosselut2018simulating}
Antoine Bosselut, Omer Levy, Ari Holtzman, Corin Ennis, Dieter Fox, and Yejin
  Choi. 2018.
\newblock Simulating action dynamics with neural process networks.
\newblock In \emph{International Conference on Learning Representations}.

\bibitem[{Chen et~al.(2021)Chen, Gan, Cheng, Zhou, Xiao, and
  Li}]{chen2021unsupervised}
Jiangjie Chen, Chun Gan, Sijie Cheng, Hao Zhou, Yanghua Xiao, and Lei Li. 2021.
\newblock Unsupervised editing for counterfactual stories.
\newblock \emph{arXiv preprint arXiv:2112.05417}.

\bibitem[{Chomsky(1956)}]{chomsky1956syntactic}
Noam Chomsky. 1956.
\newblock Syntactic structures.

\bibitem[{Donatelli et~al.(2021)Donatelli, Schmidt, Biswas, K{\"o}hn, Zhai, and
  Koller}]{donatelli2021aligning}
Lucia Donatelli, Theresa Schmidt, Debanjali Biswas, Arne K{\"o}hn, Fangzhou
  Zhai, and Alexander Koller. 2021.
\newblock Aligning actions across recipe graphs.
\newblock In \emph{Proceedings of the 2021 Conference on Empirical Methods in
  Natural Language Processing}, pages 6930--6942.

\bibitem[{Fang et~al.(2022)Fang, Baldwin, and Verspoor}]{fang2022does}
Biaoyan Fang, Timothy Baldwin, and Karin Verspoor. 2022.
\newblock What does it take to bake a cake? the reciperef corpus and anaphora
  resolution in procedural text.
\newblock In \emph{Findings of the Association for Computational Linguistics:
  ACL 2022}, pages 3481--3495.

\bibitem[{Gibson and Levin(1975)}]{gibson1975psychology}
Eleanor~J Gibson and Harry Levin. 1975.
\newblock \emph{The psychology of reading.}
\newblock The MIT press.

\bibitem[{H.~Lee et~al.(2020)H.~Lee, Shu, Achananuparp, Prasetyo, Liu, Lim, and
  Varshney}]{h2020recipegpt}
Helena H.~Lee, Ke~Shu, Palakorn Achananuparp, Philips~Kokoh Prasetyo, Yue Liu,
  Ee-Peng Lim, and Lav~R Varshney. 2020.
\newblock Recipegpt: Generative pre-training based cooking recipe generation
  and evaluation system.
\newblock In \emph{Companion Proceedings of the Web Conference 2020}, pages
  181--184.

\bibitem[{Huang(1998)}]{huang1998logical}
C-T~James Huang. 1998.
\newblock \emph{Logical relations in Chinese and the theory of grammar}.
\newblock Taylor \& Francis.

\bibitem[{Johnson et~al.(2017)Johnson, Hariharan, Van Der~Maaten, Fei-Fei,
  Lawrence~Zitnick, and Girshick}]{johnson2017clevr}
Justin Johnson, Bharath Hariharan, Laurens Van Der~Maaten, Li~Fei-Fei,
  C~Lawrence~Zitnick, and Ross Girshick. 2017.
\newblock Clevr: A diagnostic dataset for compositional language and elementary
  visual reasoning.
\newblock In \emph{Proceedings of the IEEE conference on computer vision and
  pattern recognition}, pages 2901--2910.

\bibitem[{Keysers et~al.(2019)Keysers, Sch{\"a}rli, Scales, Buisman, Furrer,
  Kashubin, Momchev, Sinopalnikov, Stafiniak, Tihon
  et~al.}]{keysers2019measuring}
Daniel Keysers, Nathanael Sch{\"a}rli, Nathan Scales, Hylke Buisman, Daniel
  Furrer, Sergii Kashubin, Nikola Momchev, Danila Sinopalnikov, Lukasz
  Stafiniak, Tibor Tihon, et~al. 2019.
\newblock Measuring compositional generalization: A comprehensive method on
  realistic data.
\newblock In \emph{International Conference on Learning Representations}.

\bibitem[{Kiddon et~al.(2015)Kiddon, Ponnuraj, Zettlemoyer, and
  Choi}]{kiddon2015mise}
Chlo{\'e} Kiddon, Ganesa~Thandavam Ponnuraj, Luke Zettlemoyer, and Yejin Choi.
  2015.
\newblock Mise en place: Unsupervised interpretation of instructional recipes.
\newblock In \emph{Proceedings of the 2015 Conference on Empirical Methods in
  Natural Language Processing}, pages 982--992.

\bibitem[{Kiddon et~al.(2016)Kiddon, Zettlemoyer, and
  Choi}]{kiddon2016globally}
Chlo{\'e} Kiddon, Luke Zettlemoyer, and Yejin Choi. 2016.
\newblock Globally coherent text generation with neural checklist models.
\newblock In \emph{Proceedings of the 2016 conference on empirical methods in
  natural language processing}, pages 329--339.

\bibitem[{Kim and Linzen(2020)}]{kim2020cogs}
Najoung Kim and Tal Linzen. 2020.
\newblock Cogs: A compositional generalization challenge based on semantic
  interpretation.
\newblock In \emph{Proceedings of the 2020 Conference on Empirical Methods in
  Natural Language Processing (EMNLP)}, pages 9087--9105.

\bibitem[{Kiritchenko and Mohammad(2017)}]{kiritchenko2017best}
Svetlana Kiritchenko and Saif Mohammad. 2017.
\newblock Best-worst scaling more reliable than rating scales: A case study on
  sentiment intensity annotation.
\newblock In \emph{Proceedings of the 55th Annual Meeting of the Association
  for Computational Linguistics (Volume 2: Short Papers)}, pages 465--470.

\bibitem[{Lake and Baroni(2018)}]{lake2018generalization}
Brenden Lake and Marco Baroni. 2018.
\newblock Generalization without systematicity: On the compositional skills of
  sequence-to-sequence recurrent networks.
\newblock In \emph{International conference on machine learning}, pages
  2873--2882. PMLR.

\bibitem[{Lake(2019)}]{lake2019compositional}
Brenden~M Lake. 2019.
\newblock Compositional generalization through meta sequence-to-sequence
  learning.
\newblock \emph{Advances in neural information processing systems}, 32.

\bibitem[{Li et~al.(2021)Li, Li, Ni, and McAuley}]{li2021share}
Shuyang Li, Yufei Li, Jianmo Ni, and Julian McAuley. 2021.
\newblock Share: a system for hierarchical assistive recipe editing.
\newblock \emph{arXiv preprint arXiv:2105.08185}.

\bibitem[{Louviere and Woodworth(1991)}]{louviere1991best}
Jordan~J Louviere and George~G Woodworth. 1991.
\newblock Best-worst scaling: A model for the largest difference judgments.
\newblock Technical report, Working paper.

\bibitem[{MacQueen et~al.(1967)}]{macqueen1967some}
James MacQueen et~al. 1967.
\newblock Some methods for classification and analysis of multivariate
  observations.
\newblock In \emph{Proceedings of the fifth Berkeley symposium on mathematical
  statistics and probability}, volume~1, pages 281--297. Oakland, CA, USA.

\bibitem[{Majumder et~al.(2019)Majumder, Li, Ni, and
  McAuley}]{majumder2019generating}
Bodhisattwa~Prasad Majumder, Shuyang Li, Jianmo Ni, and Julian McAuley. 2019.
\newblock Generating personalized recipes from historical user preferences.
\newblock In \emph{Proceedings of the 2019 Conference on Empirical Methods in
  Natural Language Processing and the 9th International Joint Conference on
  Natural Language Processing (EMNLP-IJCNLP)}, pages 5976--5982.

\bibitem[{Mori et~al.(2014)Mori, Maeta, Yamakata, and Sasada}]{mori2014flow}
Shinsuke Mori, Hirokuni Maeta, Yoko Yamakata, and Tetsuro Sasada. 2014.
\newblock Flow graph corpus from recipe texts.
\newblock In \emph{Proceedings of the Ninth International Conference on
  Language Resources and Evaluation (LREC'14)}, pages 2370--2377.

\bibitem[{Nye et~al.(2020)Nye, Solar-Lezama, Tenenbaum, and
  Lake}]{nye2020learning}
Maxwell Nye, Armando Solar-Lezama, Josh Tenenbaum, and Brenden~M Lake. 2020.
\newblock Learning compositional rules via neural program synthesis.
\newblock \emph{Advances in Neural Information Processing Systems},
  33:10832--10842.

\bibitem[{Pan et~al.(2020)Pan, Chen, Wu, Liu, Ngo, Kan, Jiang, and
  Chua}]{pan2020multi}
Liang-Ming Pan, Jingjing Chen, Jianlong Wu, Shaoteng Liu, Chong-Wah Ngo,
  Min-Yen Kan, Yugang Jiang, and Tat-Seng Chua. 2020.
\newblock Multi-modal cooking workflow construction for food recipes.
\newblock In \emph{Proceedings of the 28th ACM International Conference on
  Multimedia}, pages 1132--1141.

\bibitem[{Papineni et~al.(2002)Papineni, Roukos, Ward, and
  Zhu}]{papineni2002bleu}
Kishore Papineni, Salim Roukos, Todd Ward, and Wei-Jing Zhu. 2002.
\newblock Bleu: a method for automatic evaluation of machine translation.
\newblock In \emph{Proceedings of the 40th annual meeting of the Association
  for Computational Linguistics}, pages 311--318.

\bibitem[{Qin et~al.(2019)Qin, Bosselut, Holtzman, Bhagavatula, Clark, and
  Choi}]{qin2019counterfactual}
Lianhui Qin, Antoine Bosselut, Ari Holtzman, Chandra Bhagavatula, Elizabeth
  Clark, and Yejin Choi. 2019.
\newblock Counterfactual story reasoning and generation.
\newblock In \emph{Proceedings of the 2019 Conference on Empirical Methods in
  Natural Language Processing and the 9th International Joint Conference on
  Natural Language Processing (EMNLP-IJCNLP)}, pages 5043--5053.

\bibitem[{Qin et~al.(2020)Qin, Shwartz, West, Bhagavatula, Hwang, Le~Bras,
  Bosselut, and Choi}]{qin2020back}
Lianhui Qin, Vered Shwartz, Peter West, Chandra Bhagavatula, Jena~D Hwang,
  Ronan Le~Bras, Antoine Bosselut, and Yejin Choi. 2020.
\newblock Back to the future: Unsupervised backprop-based decoding for
  counterfactual and abductive commonsense reasoning.
\newblock In \emph{Proceedings of the 2020 Conference on Empirical Methods in
  Natural Language Processing (EMNLP)}, pages 794--805.

\bibitem[{Radford et~al.(2019)Radford, Wu, Child, Luan, Amodei, Sutskever
  et~al.}]{radford2019language}
Alec Radford, Jeffrey Wu, Rewon Child, David Luan, Dario Amodei, Ilya
  Sutskever, et~al. 2019.
\newblock Language models are unsupervised multitask learners.
\newblock \emph{OpenAI blog}, 1(8):9.

\bibitem[{Rosenbaum and Rubin(1983)}]{rosenbaum1983central}
Paul~R Rosenbaum and Donald~B Rubin. 1983.
\newblock The central role of the propensity score in observational studies for
  causal effects.
\newblock \emph{Biometrika}, 70(1):41--55.

\bibitem[{Ruis et~al.(2020)Ruis, Andreas, Baroni, Bouchacourt, and
  Lake}]{ruis2020benchmark}
Laura Ruis, Jacob Andreas, Marco Baroni, Diane Bouchacourt, and Brenden~M Lake.
  2020.
\newblock A benchmark for systematic generalization in grounded language
  understanding.
\newblock \emph{Advances in neural information processing systems},
  33:19861--19872.

\bibitem[{Sakib et~al.(2021)Sakib, Baez, Paulius, and
  Sun}]{sakib2021evaluating}
Md~Sadman Sakib, Hailey Baez, David Paulius, and Yu~Sun. 2021.
\newblock Evaluating recipes generated from functional object-oriented network.
\newblock \emph{arXiv preprint arXiv:2106.00728}.

\bibitem[{Shaw et~al.(2021)Shaw, Chang, Pasupat, and
  Toutanova}]{shaw2021compositional}
Peter Shaw, Ming-Wei Chang, Panupong Pasupat, and Kristina Toutanova. 2021.
\newblock Compositional generalization and natural language variation: Can a
  semantic parsing approach handle both?
\newblock In \emph{Proceedings of the 59th Annual Meeting of the Association
  for Computational Linguistics and the 11th International Joint Conference on
  Natural Language Processing (Volume 1: Long Papers)}, pages 922--938.

\bibitem[{Wei{\ss}enhorn et~al.(2022)Wei{\ss}enhorn, Donatelli, and
  Koller}]{weissenhorn2022compositional}
Pia Wei{\ss}enhorn, Lucia Donatelli, and Alexander Koller. 2022.
\newblock Compositional generalization with a broad-coverage semantic parser.
\newblock In \emph{Proceedings of the 11th Joint Conference on Lexical and
  Computational Semantics}, pages 44--54.

\bibitem[{Xu et~al.(2020)Xu, Zhang, and Dong}]{xu2020cluecorpus2020}
Liang Xu, Xuanwei Zhang, and Qianqian Dong. 2020.
\newblock Cluecorpus2020: A large-scale chinese corpus for pre-training
  language model.
\newblock \emph{arXiv preprint arXiv:2003.01355}.

\bibitem[{Yin et~al.(2021)Yin, Li, Hu, Peng, and Chang}]{yin2021broaden}
Da~Yin, Liunian~Harold Li, Ziniu Hu, Nanyun Peng, and Kai-Wei Chang. 2021.
\newblock \href {https://arxiv.org/abs/2109.06860} {{Broaden the Vision:
  Geo-Diverse Visual Commonsense Reasoning}}.
\newblock In \emph{EMNLP}.

\bibitem[{Zhang et~al.(2020)Zhang, Kishore, Wu, Weinberger, and
  Artzi}]{bert-score}
Tianyi Zhang, Varsha Kishore, Felix Wu, Kilian~Q. Weinberger, and Yoav Artzi.
  2020.
\newblock \href {https://openreview.net/forum?id=SkeHuCVFDr} {Bertscore:
  Evaluating text generation with bert}.
\newblock In \emph{International Conference on Learning Representations}.

\bibitem[{Zhao et~al.(2019)Zhao, Chen, Zhang, Zhao, Liu, Lu, Chen, Deng, Ju,
  and Du}]{zhao2019uer}
Zhe Zhao, Hui Chen, Jinbin Zhang, Xin Zhao, Tao Liu, Wei Lu, Xi~Chen, Haotang
  Deng, Qi~Ju, and Xiaoyong Du. 2019.
\newblock Uer: An open-source toolkit for pre-training models.
\newblock \emph{EMNLP-IJCNLP 2019}, page 241.

\end{thebibliography}
\bibliographystyle{acl_natbib}

\clearpage
\appendix

\section*{Appendix}
\begin{table*}[t]
    \centering
    \small
    \begin{tabular}{lcccccccccc}
    \toprule
    & \multicolumn{5}{c}{\textbf{Hard}} & \multicolumn{5}{c}{\textbf{Soft}}\\
    \cmidrule(lr){2-6}
    \cmidrule(lr){7-11} 
    & \textbf{P} & \textbf{R} & $\mathbf{F_1}$ & $\mathbf{F_1^+}$ & $\mathbf{F_1^-}$ & \textbf{P} & \textbf{R} & $\mathbf{F_1}$ & $\mathbf{F_1^+}$ & $\mathbf{F_1^-}$\\
    \midrule
    GPT-2 (D) & 11.8 & 33.8 & 15.9 & 13.3 & 26.7 & 21.7 & \textbf{56.8} & \textbf{29.0} & \textbf{27.7} & 31.0 \\
    GPT-2 (D+R) & 13.1 & 34.3 & 17.2 & 15.3 & 26.9 & 21.4 & 53.2 & 27.9 & 25.8 & 31.2 \\
    \textsc{EduCat} & 8.9 & 16.2 & 10.0 & 4.4 & 23.2 & 17.7 & 29.1 & 19.5 & 11.5 & 27.8 \\
    \textsc{Delorean} & \textbf{13.5} & \textbf{34.6} & \textbf{17.5} & \textbf{15.4} & \textbf{27.3} & \textbf{22.4} & 52.5 & 28.6 & 26.5 & \textbf{31.6} \\
    \bottomrule
    \end{tabular}
    \caption{Full results of action-level evaluation. $F_1^+$ and $F_1^-$ indicate $F_1$ on actions to insert and actions to remove respectively.}
    \label{table-eval-appendix}
\end{table*}
\begin{table*}[ht]
    \centering
    \small
    \renewcommand{\arraystretch}{1.2}
    \begin{tabularx}{\textwidth}{X}
    \toprule
    \textbf{Base Dish: steamed turbot, Target Dish: steamed hairy crab} \\
    \midrule
    \textbf{Base recipe:} Cut 1/3 of the green onion and ginger into fine julienne strips, ... (seasoning preparation) Clean the turbot, make diamond-shaped knives on both sides and sprinkle the cooking wine evenly over the fish ...\\
    \textbf{Counterfactual recipe [\textsc{EduCat}]:} Cut 1/3 of the green onion and ginger into fine julienne strips, ... (seasoning preparation) Clean the \purple{turbot}, \purple{make a few cuts on both sides} and rub with cooking wine ...\\
    \textbf{Counterfactual recipe [\textsc{Delorean}]:} \purple{After steaming, lift the crab shell, remove the gills, cut the crab body into two halves and arrange on the plate.} Add appropriate amount of oil to the pot. When the oil is hot, add onion, ginger and garlic to burst. Add appropriate amount of steamed fish soy sauce, then add appropriate amount of water. Boil the water and pour it over the crab ...\\
    \textbf{Counterfactual recipe [Expert]:} Cut 1/3 of the green onion and ginger into fine julienne strips, ... (seasoning preparation) Clean the \blue{hairy crab} and sprinkle the cooking wine evenly over the hairy crab ...\\
    \midrule
    \textbf{Base Dish: spicy stir-fried squid, Target Dish: spicy stir-fried snails} \\
    \midrule
    \textbf{Base recipe:} The squid must be blanched through water to remove the fishy. Wash the squid neck, remove the middle ink sac and squid mouth, and cut into pieces. Blanch and drain ...\\
    \textbf{Counterfactual recipe [\textsc{EduCat}]:} \purple{Wash off the sediment in squid, remove the outer skin and guts, cut into circles.} Burst large onion (cut into sections) and ginger slices with flavor ...\\
    \textbf{Counterfactual recipe [\textsc{Delorean}]:} Wash the snail, cut off the tail, soak it in salt water for half an hour, and then rinse it well with water. Pour oil in a pot, add ginger, garlic, dried chili, Sichuan pepper, star anise, cinnamon, and allspice, and stir-fry them. Add the snails and stir-fry ...\\
    \textbf{Counterfactual recipe [Expert]:} Clean the snails, soak the snails in lightly salted water for 1 hour to \blue{spit out the sediment}. Boil water in a pot, fish out the soaked snails into the boiling water and \blue{blanch} them quickly. Fish them out and drain them ...\\
    \bottomrule
    \end{tabularx}
    \caption{Examples of generated recipes of two dish pairs.}
    \label{table-case}
\end{table*}
\section{Implementation Details}
\subsection{Dish Pairs Selection}
\label{appendix:dish_pairs}
\begin{table*}[t]
    \centering
    \small
    \renewcommand{\arraystretch}{1.2}
    \setlength{\tabcolsep}{1pt}
    \begin{CJK*}{UTF8}{gbsn}
    \begin{tabular}{llll}
    \toprule
    \textbf{Base Dish} & & \textbf{Target Dish}& \\
    \midrule
    香辣小龙虾 & Spicy Crayfish & 香辣鸡爪 & Spicy Chicken Feet\\
    宫保鸡丁 & Kung Pao Chicken & 宫保虾球 & Kung Pao Shrimp Balls \\
    清炒小白菜 & Stir-fried Baby Cabbage & 清炒丝瓜 & Stir-fried Loofah \\
    白灼菜心 & Blanched Choy Sum & 白灼秋葵 & Blanched Okra \\
    酸辣鸡爪 & Sour-and-hot Chicken Feet & 酸辣鸡杂 & Sour-and-hot Chicken Offal \\
    辣炒年糕 & Spicy Stir-fried Rice Cakes & 辣炒花甲 & Spicy Stir-fried Clams \\
    焖牛腩 & Stewed Beef Brisket & 焖猪脚 & Stewed Pork Feet \\
    醋溜白菜 & Chinese Cabbage with Vinegar Sauce & 醋溜土豆丝 & Shredded Potatoes with Vinegar Sauce \\
    烤鸡翅 & Grilled Chicken Wings & 烤羊排 & Grilled Lamb Chops \\
    清蒸多宝鱼 & Steamed Turbot & 清蒸大闸蟹 & Steamed Hairy Crab \\
    麻辣豆腐 & Spicy Tofu & 麻辣土豆片 & Spicy Potato Chips \\
    炒花蛤 & Stir-Fried Clams & 炒蚬子 & Stir-Fried Razor Fish \\
    凉拌苦瓜 & Cold Bitter Gourd & 凉拌皮蛋 & Cold Century Egg \\
    爆炒牛肉 & Sautéed Beef & 爆炒鸡胗 & Sautéed Chicken Gizzards \\
    炖蛋 & Braised Egg & 炖燕窝 & Braised Bird’s Nest \\
    烧牛腩 & Braised Beef Brisket  & 烧小黄鱼 & Braised Small Yellow Croaker \\
    麻辣牛肉干 & Spicy Beef Jerky & 麻辣香肠 & Spicy Sausage \\
    煎口蘑 & Pan-fried Mushrooms & 煎年糕 & Pan-fried Rice Cakes \\
    酸辣藕片 & Sour-and-hot Lotus Roots & 酸辣包菜 & Sour-and-hot Cabbage \\
    酸辣土豆丝 & Spicy Shredded Potatoes in Vinegar & 酸辣蕨根粉 & Spicy Fern Root Noodles in Vinegar \\
    干煸鸡 & Sautéed Chicken & 干煸肥肠 & Sautéed Pork Intestines \\
    辣炒鱿鱼 & Spicy Stir-fried Squid & 辣炒田螺 & Spicy Stir-fried Snails \\
    凉拌苦菊 & Cold Bitter Chrysanthemum & 凉拌海蜇 & Cold Jellyfish \\
    炒蒜苗 & Stir-Fried Garlic Sprouts & 炒藕片 & Stir-Fried Lotus Root Slices \\
    油焖大虾 & Braised Prawn & 油焖尖椒 & Braised Hot Pepper \\
    拔丝地瓜 & Sweet Potatoes in Hot Toffee & 拔丝苹果 & Apple in Hot Toffee \\
    煎鱼 & Pan-fried Fish & 煎猪扒 & Pan-fried Pork Chops \\
    酿苦瓜 & Stuffed Bitter Melon with Pork and Shrimp & 酿茄子 & Stuffed Eggplant with Pork and Shrimp \\
    \multirow{2}{*}{溜鱼片} & \multirow{2}{*}{Quick-Fried Sliced Fish with Brown Sauce} & \multirow{2}{*}{溜肥肠} & Quick-Fried Pork Intestines \\ [-0.3em]
    & & & \quad with Brown Sauce \\
    脆皮豆腐 & Crispy Tofu & 脆皮烧肉 & Crispy Roast Pork \\
    白菜炖豆腐 & Braised Chinese Cabbage with Tofu & 白菜炖粉条 & Braised Chinese Cabbage with Vermicelli \\
    辣白菜炒五花肉 & Fried Pork with Spicy Cabbage & 辣白菜炒年糕 & Fried Rice Cakes with Spicy Cabbage \\
    冰糖雪梨 & White Fungus with Rock Sugar & 冰糖金桔 & Kumquat with Rock Sugar \\
    蚝油生菜 & Sautéed Lettuce with Oyster Sauce & 蚝油西兰花 & Sautéed Broccoli with Oyster Sauce \\
    泡椒凤爪 & Chicken Feet with Pickled Peppers & 泡椒牛蛙 & Bullfrog with Pickled Peppers \\
    \multirow{2}{*}{蒜蓉金针菇} & Sautéed Needle Mushroom & \multirow{2}{*}{蒜蓉油麦菜} & \multirow{2}{*}{Sautéed Leaf Lettuce with Mashed Garlic}\\ [-0.3em]
    & \quad with Mashed Garlic & & \\
    芝士焗大虾 & Baked Prawns with Cheese & 芝士焗土豆泥 & Baked Mashed Potatoes with Cheese \\
    \multirow{2}{*}{酒酿蛋} & \multirow{2}{*}{Egg in Fermented Rice Wine} & \multirow{2}{*}{酒酿圆子} & Glutinous Rice Balls \\ [-0.3em]
    & & & \quad in Fermented Rice Wine \\
    茭白炒肉 & Stir-fried Pork with Wild Rice & 蒜苗炒肉 & Stir-fried Pork with Garlic Sprouts \\
    芦笋炒虾仁 & Sautéed Shrimp with Green Asparagus & 芦笋炒蘑菇 & Sautéed Mushroom with Green Asparagus \\
    包菜炒粉丝 & Scrambled Cabbage with Vermicelli & 鸡蛋炒粉丝 & Scrambled Eggs with Vermicelli \\
    羊肉炖萝卜 & Stewed Lamb with Turnip & 牛腩炖萝卜 & Stewed Beef Brisket with Turnip \\
    木瓜炖牛奶 & Stewed Papaya in Milk & 花胶炖牛奶 & Stewed Fish Maw in Milk \\
    青椒塞肉 & Green Pepper with Pork Stuffing & 油面筋塞肉 & Wheat Gluten with Pork Stuffing \\
    排骨豆角焖面 & Stewed Noodles with Pork Ribs and Beans & 土豆豆角焖面 & Stewed Noodles with Potatoes and Beans \\
    凉拌豆腐 & Cold Tofu & 凉拌皮蛋豆腐 & Cold Tofu with Century Egg \\
    炒牛肉 & Fried Beef & 胡萝卜炒牛肉 & Fried Beef with Carrot \\
    \multirow{2}{*}{肉末四季豆} & \multirow{2}{*}{Stir-fried Green Beans with Minced Pork} & \multirow{2}{*}{榄菜肉末四季豆} & Stir-fried Green Beans with Minced Pork\\ [-0.3em]
    & & & \quad and Olive Vegetables \\
    焖鸡 & Smothered Chicken & 鲍鱼焖鸡 & Smothered Chicken with Abalone \\
    \multirow{2}{*}{蒜蓉粉丝娃娃菜} & Steam Baby Cabbage with Vermicelli & \multirow{2}{*}{蒜蓉粉丝扇贝} & \multirow{2}{*}{Steam Scallops with Vermicelli and Garlic} \\[-0.3em]
    & \quad and Garlic & & \\
    \bottomrule
    \end{tabular}
    \end{CJK*}
    \caption{List of dish pairs used in evaluation.}
    \label{table-dish-full}
\end{table*}
We provide the full list of dish pairs in Table~\ref{table-dish-full}. The 50 dish pairs involve appetizers, main dishes, soups, staples, and desserts. Both the translated English version and the original Chinese version are provided.

We choose the dish pairs based on the following criteria:

1. The dishes are common in Chinese cuisine. Each dish has at least 50 recipes in our \textsc{XiaChuFang} dataset. 

2. Recipes of the dishes have not been seen by the PLMs. We use CLUECorpus2020, an open-sourced large-scale Chinese corpus, as the pretraining corpus of PLMs, and verify that none of the recipes for these dishes are in the corpus.

3. Models have the opportunity to learn about the ingredients and flavors. We ensure that for each ingredient or flavor appearing in the dish pairs, there are at least 5 other dishes in \textsc{XiaChuFang} that share the same ingredient or flavor. 

\subsection{Recipe Parsing}
\label{appendix:parsing}
As we did not find off-the-shelf Chinese recipe parsing models, we parse recipes from scratch. The parsing process converts a recipe into a list of clustered actions.

\paragraph{Glossary} We build a glossary of common verbs, ingredients, and tools; and cluster them into 92 verb classes $V$, 476 ingredient classes $I$, and 20 tool classes $T$. This is to tackle the \emph{multiple surface forms} of the same or similar meaning.
For example, \{\begin{CJK*}{UTF8}{gbsn}
加入, 放入, 添加, 加, 下, ...
\end{CJK*}\}
all mean \emph{add} in Chinese recipes.
The classes are clustered by K-means~\citep{macqueen1967some} of word embeddings and adjusted by experts.

\paragraph{Dependency Parsing} We perform dependency parsing on recipes, and build actions from verbs in the dependency parsing tree. If a verb has noun children that are ingredients or cooking tools, they are also included in the action.
For example, as shown in Figure~\ref{fig-pipeline}, the phrase \emph{cut off the head} is parsed into the action $(cut, head, \varnothing)$.
As we aim to collect common patterns, we only consider words in the glossary and filter out the rare ones.  

\paragraph{Proto-actions} The actions are grouped by word classes: if two actions' verbs are in one verb class, and the classes of their ingredients and tools are the same, they are clustered into one proto-action
\begin{equation}
    a=(v, igs, tools),
\end{equation}
where $v \in V$, $igs \subseteq I$, and $tools \subseteq T$.
Therefore, a recipe is converted into a list of proto-actions. A recipe is made up of 19.6 proto-actions on average.

\paragraph{Limitation} The parsing is quite coarse as we do not have fine-grained parsed Chinese recipes for model training. For example, we do not consider the attribute knowledge: like the cooking temperature and the amount of ingredients. We plan to annotate recipes and build accurate parsing tools in the future. 

\subsection{Annotation Samples}
\label{appendix:annotation}
\begin{figure*}[ht]
    \centering
    \begin{subfigure}[t]{0.48\textwidth}
        \centering
        \includegraphics[width=\columnwidth]{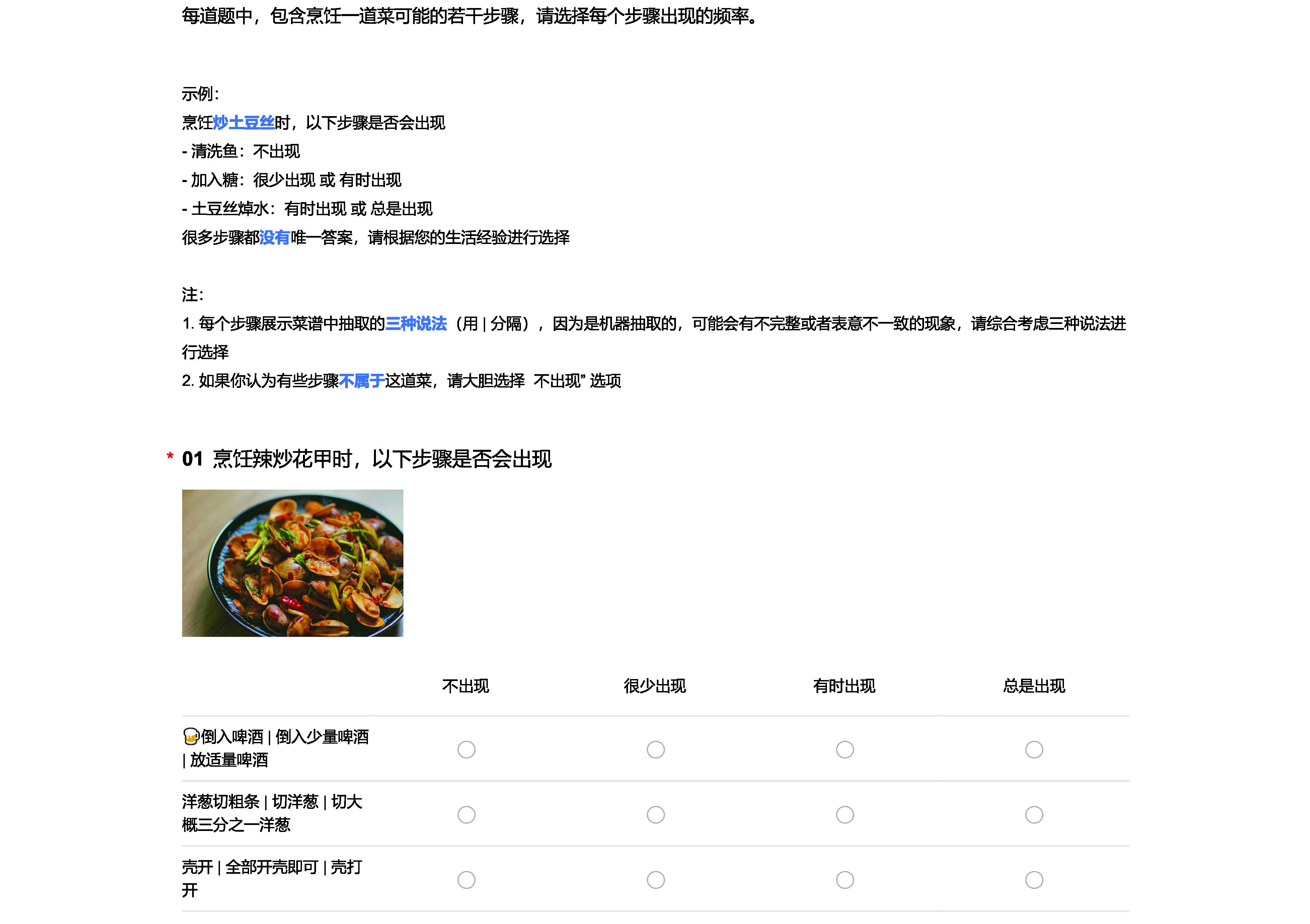}
        \caption{Chinese version.}
    \end{subfigure}%
    ~ 
    \begin{subfigure}[t]{0.48\textwidth}
        \centering
        \includegraphics[width=\columnwidth]{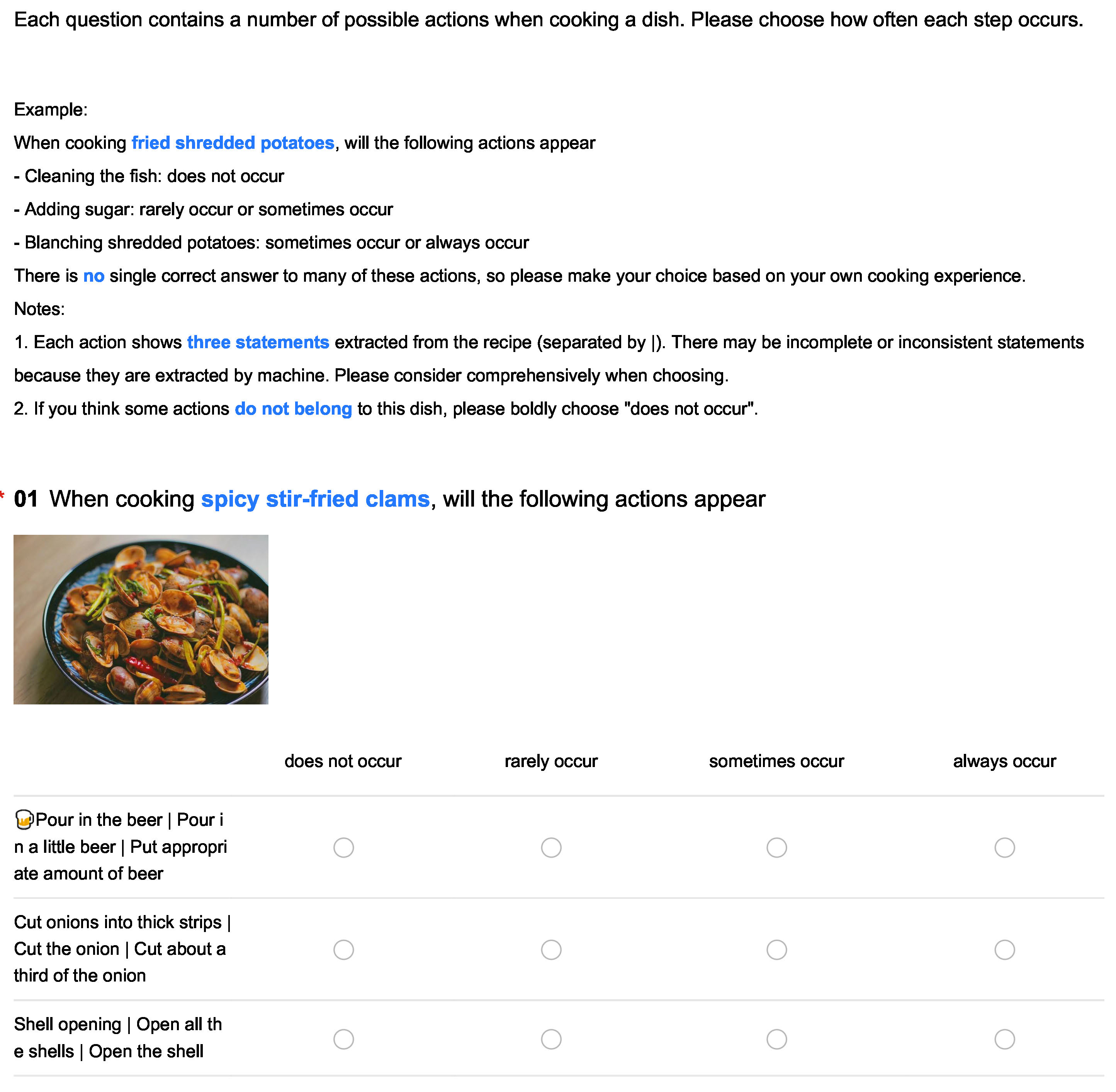}
        \caption{Translated English version.}
    \end{subfigure}
    \caption{Annotation sample of pivot action collection.}
    \label{fig-annotation1}
\end{figure*}
\begin{figure*}[!ht]
    \centering
    \begin{subfigure}[t]{0.48\textwidth}
        \centering
        \includegraphics[width=\columnwidth]{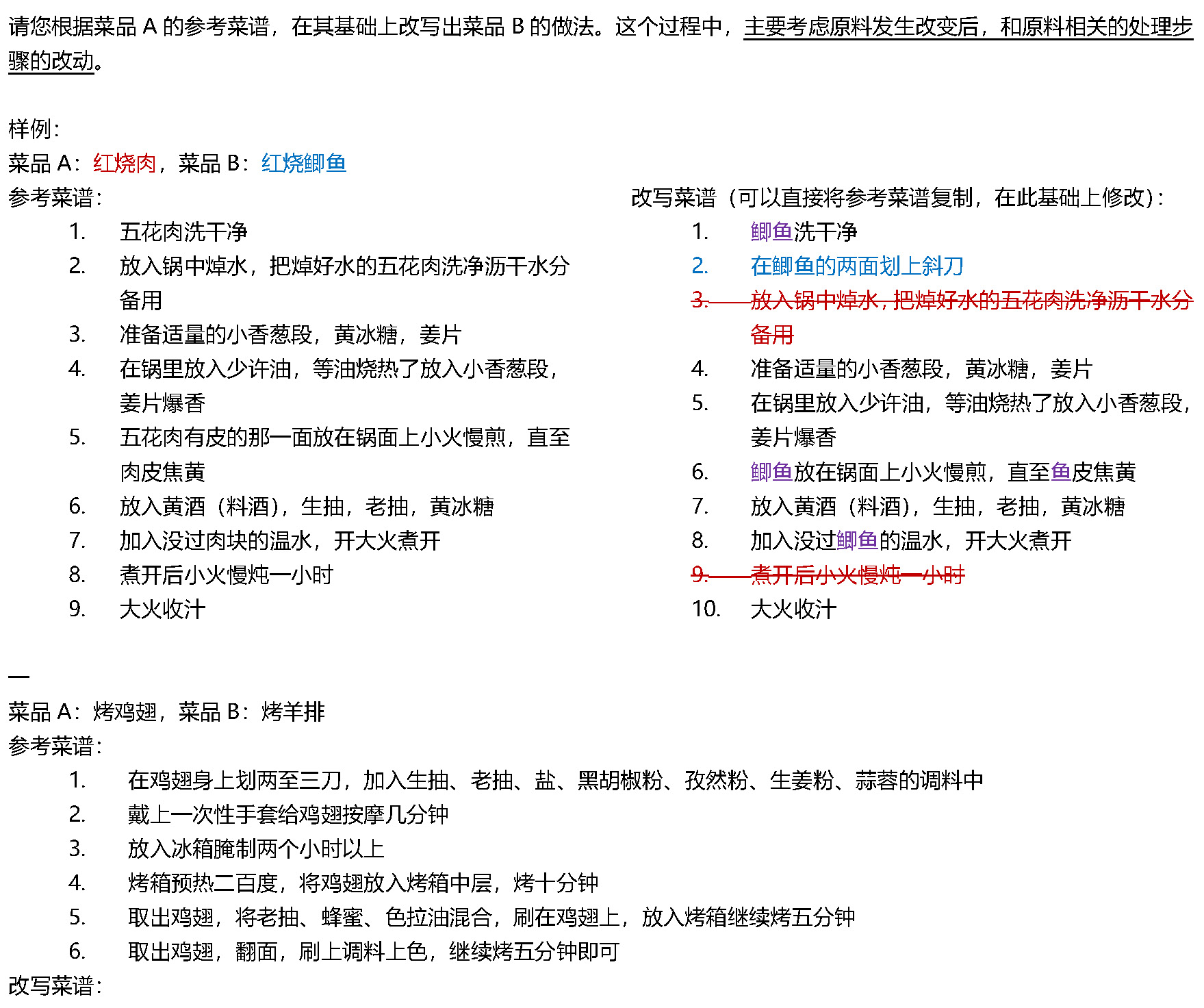}
        \caption{Chinese version.}
    \end{subfigure}%
    ~ 
    \begin{subfigure}[t]{0.48\textwidth}
        \centering
        \includegraphics[width=\columnwidth]{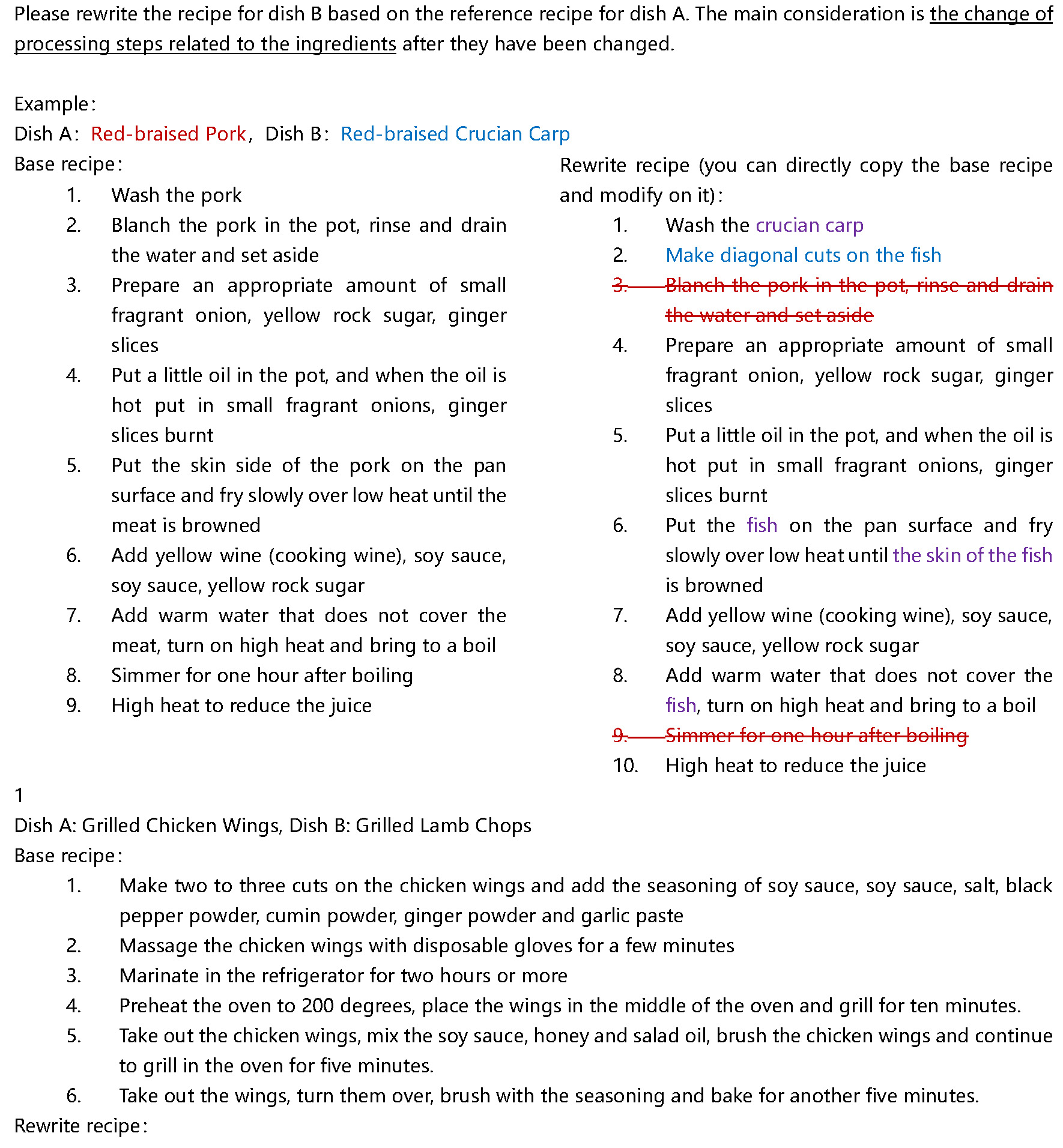}
        \caption{Translated English version.}
    \end{subfigure}
    \caption{Annotation sample of manually counterfactual writing.}
    \label{fig-annotation2}
\end{figure*}
We demonstrate an annotation sample of pivot action collection in Figure~\ref{fig-annotation1}, and a counterfactual writing sample in Figure~\ref{fig-annotation2}. 

\subsection{Computing Infrastructure}
Experiments are conducted on NVIDIA GeForce RTX 3090 GPU. It takes \textsc{EduCat} two minutes to generate a recipe on average, and no more than half a minute for other models. The parameter size of the GPT-2 we used is 102M.

\section{Additional Results}
\subsection{Disagreement Analysis of Pilot Study}
\label{appendix:disagreement}
We examine the 31 disagreement actions in $P_R$ and $P_I$ and find that the main reason is that the cooking process is subjective, especially in informal circumstances like home cooking. People will change some auxiliary ingredients or steps according to their personal preferences, but the principal steps and ingredients are highly consistent. For example, some people add sugar when cooking spicy stir-fried squid, while some never do it. In the 31 disagreements, 68\% are related to the auxiliary ingredients. We think the disagreements would not affect our main findings as we care more about the steps related to principal ingredients. 

\subsection{Action-level Evaluation Results}
We do not place the $\mathbf{F_1^+}$ and $\mathbf{F_1^-}$ of hard action-level evaluation in Table~\ref{table-action} due to space limit. The full results are provided in Table~\ref{table-eval-appendix}.

\section{Case Study}
\label{appendix:case}
We demonstrate generated recipes of two dish pairs in Table~\ref{table-case} to analyze some common mistakes of counterfactual generation models. We only show part of the text for the ease of reading, and the full text are in Table~\ref{table-case-full-1} and Table~\ref{table-case-full-2}.

\textbf{Failing to modify the ingredient.} \textsc{EduCat} has difficulty in generating the newly added ingredient and removing the replaced one. In the case of \emph{steamed hairy crab}, it follows the base recipe to \emph{clean the turbot}, overlooking the new ingredient \emph{hairy crab} in the target dish; and it still uses the ingredient \emph{squid} in the base dish when cooking \emph{spicy stir-fried snails}. These illustrate the instability of \textsc{EduCat} in surface-level composition. 

\textbf{Failing to remove irrelevant actions.} In the case of \emph{steamed hairy crab}, \textsc{EduCat} keeps the action \emph{make a few cuts on both sides}, which is not appropriate for a hairy crab as it has a hard shell; and in the case of \emph{spicy stir-fried snails}, it does not delete \emph{remove the outer skin}, but snails have hard shells. These show that the model does not learn the relationship between the actions and ingredients in the finetuning stage.

\textbf{Failing to generate necessary actions.} When cooking snails, we should let them \emph{spit out the sediment} before putting them in the pot, but \textsc{Delorean} does not generate the action. The action \emph{blanch} can remove the fishy smell from aquatic products, and is suitable for both squid and snails, but \textsc{Delorean} wrongly deletes it. It does not understand the purpose of these actions in this case.

\textbf{Wrong order.} \textsc{Delorean} begins the recipe of \emph{steamed hairy crab} with \emph{lift the crab shell}. This is an appropriate action for this dish, but should not happen at the beginning when the crab is raw. Models should learn the underlying causal dependency of actions to avoid this kind of mistake. 

\begin{table*}[t]
    \centering
    \small
    \renewcommand{\arraystretch}{1.2}
    \begin{CJK*}{UTF8}{gbsn}
    \begin{tabularx}{\textwidth}{X}
    \toprule
    \textbf{Base Dish: steamed turbot 清蒸多宝鱼, Target Dish: steamed hairy crab 清蒸大闸蟹} \\
    \midrule
    \textbf{Base recipe:} Cut 1/3 of the green onion and ginger into fine julienne strips, cut the rest of the green onion into inch pieces, slice the ginger, cut the red pepper into fine julienne strips, and wash the cilantro and set aside. Mix soy sauce, oyster sauce, sugar, a few drops of fish sauce and 4 tablespoons of water, boil in a small pot or microwave for 2 minutes and set aside. Clean the turbot, make diamond-shaped knives on both sides and sprinkle the cooking wine evenly over the fish. And spread the sliced green onion and ginger evenly on the top and bottom of the fish in a suitable size plate. Add to the steamer for 8-9 minutes on high heat, and then turn off the heat. Remove from the heat but do not open the lid and steam for 2 minutes. Remove the onion and ginger from the steamed fish and discard the broth from the plate. Drizzle the cooked soy sauce along the side of the fish, then spread shredded green onion, ginger and red pepper over the fish, and surround the fish with cilantro. Heat 3 tablespoons of vegetable oil in a small pot until it smokes (70 or 80 percent heat) and pour it over the shredded green onion and ginger. 将1/3的葱和姜切成细丝，剩下的葱切寸段，姜切片，红辣椒切细丝，香菜洗净待用。将生抽、蚝油、糖、几滴鱼露和4大勺清水混合，用小锅煮沸或微波炉转2分钟待用。将多宝鱼处理干净，两面打菱形花刀，在鱼身上均匀的撒上料酒。并将葱段和姜片均匀铺在鱼的上下方，放在大小合适的盘中。入蒸锅大火8-9分钟即关火，离火但不要开锅盖，虚蒸2分钟。蒸好的鱼去掉葱姜，盘里的汤汁也弃去不用。沿鱼身边淋入煮好的豉油，再在鱼身上铺葱姜丝和红椒丝，香菜围边。取小锅内放3大勺植物油，烧到冒白烟的程度（7、8成热），浇在葱姜丝上即可。\\
    \textbf{Counterfactual recipe [\textsc{EduCat}]:} Cut 1/3 of the green onion and ginger into fine julienne strips, the rest of the green onion and ginger into julienne strips, the dried chili pepper into small pieces, and the cilantro into pieces. Bring soy sauce, oyster sauce, sugar, a few drops of fish sauce and a tablespoon of water to a boil in a small pot. Microwave for 2 minutes. Clean the turbot, make a few cuts on both sides and rub with cooking wine. Put it into a suitable container. Put on the steamer and turn off the heat for 8-9 minutes. After turning off the heat, do not open the lid and simmer for 3 minutes. Remove the ginger from the steamed fish and pour off the steaming broth without it. Pour the cooked sauce along the side of the plate. Spread onion, ginger, chili and cilantro over the fish. Heat a tablespoon of vegetable oil in a pan until it smokes (70 or 80 percent heat). Pour over the shredded green onions and ginger. 1/3的葱和姜切成细丝，剩下的葱段和姜切丝，干辣椒切小段，香菜切段。生抽、蚝油、糖、几滴鱼露和一大勺清水用小锅煮开。微波炉叮2分钟。多宝鱼清洗干净，两面切几刀，抹上料酒。放入合适的容器。上蒸锅，8-9分钟关火。关火后不要开盖，焖3分钟。蒸好的鱼去掉姜丝，蒸出的汤汁倒掉不要。沿着盘边倒入煮好的酱汁。在鱼身上铺上葱姜辣椒和香菜。锅中放一大勺植物油，烧到冒青烟的程度（7-8成热）。浇在葱姜丝上。\\
    \textbf{Counterfactual recipe [\textsc{Delorean}]:} After steaming, lift the crab shell, remove the gills, cut the crab body into two halves and arrange on the plate. Add appropriate amount of oil to the pot. When the oil is hot, add onion, ginger and garlic to burst. Add appropriate amount of steamed fish soy sauce, then add appropriate amount of water. Boil the water and pour it over the crab. Steamed hairy crab is made as follows. Wash the hairy crabs, put them into a steamer, put an appropriate amount of water in the steamer, boil over high heat and then turn the heat to medium and steam for 15 minutes. Steamed hairy crab is made as follows. Wash the hairy crabs, put them into a steamer, put an appropriate amount of water in the steamer, boil over high heat and then turn the heat to medium and steam for 15 minutes. Steamed hairy crab is made as follows. 蒸好后，把蟹壳掀开，把腮去掉，蟹身切成两半，摆盘。锅里放入适量的油，油热后放入葱姜蒜爆香，加入适量的蒸鱼豉油，再加入适量的水，烧开后浇在蟹身上。清蒸大闸蟹的做法如下。大闸蟹洗净，放入蒸锅，蒸锅里放入适量的水，大火烧开后转中火蒸15分钟。清蒸大闸蟹的做法如下。大闸蟹洗净，放入蒸锅，蒸锅里放入适量的水，大火烧开后转中火蒸15分钟。清蒸大闸蟹的做法如下。\\
    \textbf{Counterfactual recipe [Expert]:} Cut 1/3 of the green onion and ginger into fine julienne strips, cut the rest of the green onion into inch pieces, slice the ginger, cut the red pepper into fine julienne strips, and wash the cilantro and set aside. Mix soy sauce, oyster sauce, sugar, a few drops of fish sauce and 4 tablespoons of water, boil in a small pot or microwave for 2 minutes and set aside. Clean the hairy crab and sprinkle the cooking wine evenly over the hairy crab. And spread the sliced green onion and ginger evenly on the top and bottom of the hairy crab in a suitable size plate. Put into the steamer for 10-15 minutes on high heat and turn off the heat. Remove from the heat but do not open the lid and steam for 2 minutes in vain. Make a dipping sauce with shredded ginger, vinegar, etc. Remove the onion and ginger from the steamer and discard the broth from the plate, and serve the delicious and tempting hairy crabs. 将1/3的葱和姜切成细丝，剩下的葱切寸段，姜切片，红辣椒切细丝，香菜洗净待用。将生抽、蚝油、糖、几滴鱼露和4大勺清水混合，用小锅煮沸或微波炉转2分钟待用。将大闸蟹处理干净，在大闸蟹身上均匀的撒上料酒。并将葱段和姜片均匀铺在大闸蟹的上下方，放在大小合适的盘中。入蒸锅大火10-15分钟即关火，离火但不要开锅盖，虚蒸2分钟。用姜丝、醋，等调制蘸料。蒸好的大闸蟹去掉葱姜，盘里的汤汁也弃去不用，美味诱惑的大闸蟹即可食用。\\
    \bottomrule
    \end{tabularx}
    \end{CJK*}
    \caption{Counterfactually generated \emph{steamed hairy crab} recipes based on the recipe of \emph{steamed turbot}. }
    \label{table-case-full-1}
\end{table*}
\begin{table*}[t]
    \centering
    \small
    \renewcommand{\arraystretch}{1.2}
    \begin{CJK*}{UTF8}{gbsn}
    \begin{tabularx}{\textwidth}{X}
    \toprule
    \textbf{Base Dish: spicy stir-fried squid 辣炒鱿鱼, Target Dish: spicy stir-fried snails 辣炒田螺} \\
    \midrule
    \textbf{Base recipe:} The squid must be blanched through water to remove the fishy. Wash the squid neck, remove the middle ink sac and squid mouth, and cut into pieces. Blanch and drain. Heat the oil and then add chili (can be omitted), ginger, garlic into the pot one after another. Stir-fry ingredients until half cooked (I did not add green pepper), add the right amount of salt, add the squid neck. Add cooking wine, soy sauce, LaoGanMa seasoning and cook until tasty. Plate up and eat. 鱿鱼一定要过水，去腥。洗净鱿鱼颈，去除中间墨囊及鱿鱼嘴，切成块。焯过捞起沥水。辣椒（可省）姜，蒜，热油后陆续放入锅中。配料炒至半熟（青椒等，我没加），加适量盐，放入鱿鱼颈。加料酒、生抽、老干妈调料烧至入味即可。装盘开吃。\\
    \textbf{Counterfactual recipe [\textsc{EduCat}]:} Wash off the sediment in squid, remove the outer skin and guts, cut into circles. Burst large onion (cut into sections) and ginger slices with flavor (I did not add for the sake of good looks). Add the right amount of salt and add snails to fry. Add cooking wine, soy sauce, sugar, oyster sauce, and eat. 洗去泥沙鱿鱼去掉外皮和内脏，切成圈，大葱（切段），姜切片配料爆香（为了好看我没加）加适量盐加入田螺炒。加料酒、生抽、白糖、蚝油开吃。\\
    \textbf{Counterfactual recipe [\textsc{Delorean}]:} Wash the snail, cut off the tail, soak it in salt water for half an hour, and then rinse it well with water. Pour oil in a pot, add ginger, garlic, dried chili, Sichuan pepper, star anise, cinnamon, and allspice, and stir-fry them. Add the snails and stir-fry. Add wine, light soy sauce, dark soy sauce, oyster sauce, salt, sugar, and beer, cover the pot, boil on high heat, turn to medium heat and cook for 15 minutes. Add the green pepper, absorb the sauce over high heat, and remove from the pot. 田螺洗净，剪去尾部，放盐水浸泡半小时，再用清水冲洗干净。锅中放油，放入姜片，蒜片，干辣椒，花椒，八角，桂皮，香叶，炒香。放入田螺，翻炒，加入料酒，生抽，老抽，蚝油，盐，糖，加入啤酒，盖上锅盖，大火烧开，转中火煮15分钟。加入青椒，大火收汁，出锅。\\
    \textbf{Counterfactual recipe [Expert]:} Clean the snails, soak the snails in lightly salted water for 1 hour to spit out the sediment. Boil water in a pot, fish out the soaked snails into the boiling water and blanch them quickly. Fish them out and drain them.  Heat the oil and then add chili (can be omitted), ginger, garlic into the pot one after another. Add green pepper, green garlic and other ingredients and stir-fry them until half cooked, add appropriate amount of salt and add the snails. Add cooking wine, soy sauce, LaoGanMa seasoning and cook until tasty. Plate up and eat. 清洗田螺，将田螺在淡盐水中浸泡1个小时，吐泥沙。锅中烧开水，将浸泡的田螺捞出放入开水中。快速焯水，捞出，沥水。辣椒（可省）姜，蒜，热油后陆续放入锅中。加入青椒、青蒜等配料炒至半熟，加适量盐，放入田螺。加料酒、生抽、老干妈调料烧至入味即可。装盘开吃。\\
    \bottomrule
    \end{tabularx}
    \end{CJK*}
    \caption{Counterfactually generated \emph{spicy stir-fried snails} recipes based on the recipe of \emph{spicy stir-fried squid}. }
    \label{table-case-full-2}
\end{table*}

\end{document}